\documentclass[twoside,11pt]{article}

\usepackage{jmlr2e}

\usepackage{times}
\usepackage{graphicx}
\usepackage{subcaption}
\usepackage{amsmath}
\usepackage{amsfonts}
\usepackage{amssymb}
\usepackage{mathtools}
\usepackage{enumerate}
\usepackage{lscape}
\usepackage{longtable}
\usepackage{rotating}
\usepackage{multirow}
\usepackage{color}
\usepackage{url}
\usepackage{rotating}
\usepackage{comment}
\usepackage[utf8]{inputenc}
\usepackage{algorithm}
\usepackage{algpseudocode}

\usepackage{xcolor}
\usepackage{booktabs}

\newcommand{\reffig}[1]{Figure~\ref{fig:#1}}
\newcommand{\lblfig}[1]{\label{fig:#1}}
\newcommand{\lbltbl}[1]{\label{tbl:#1}}
\newcommand{\reftbl}[1]{Table~\ref{tbl:#1}}

\usepackage{lastpage}
\jmlrheading{24}{2024}{1-\pageref{LastPage}}{1/24}{}{24-0000}{V. Kungurtsev, Y. Peng, S. Vahidian, J. Gu, S. Vahidian, A. Quinn, F. Idlahcen and Y. Chen}

\ShortHeadings{DD First Principles}{V. Kungurtsev, Y. Peng, S. Vahidian, J. Gu, S. Vahidian, A. Quinn, F. Idlahcen and Y. Chen}
\firstpageno{1}

\begin{document}


\title{Dataset Distillation from First Principles: Integrating Core Information Extraction and Purposeful Learning}

\author{ 
       \name Vyacheslav Kungurtsev \email vyacheslav.kungurtsev@fel.cvut.cz \\
       \addr Czech Technical University in Prague 
       \AND
       \name Yuanfang Peng \email  
       ypengbx@connect.ust.hk\\
       \addr Hong Kong University of Science and Technology
       \AND 
       \name Jianyang Gu \email gu.1220@osu.edu \\
       \addr Ohio State University
       \AND
       \name Saeed Vahidian \email saeed.vahidian@duke.edu\\  
       \addr Duke University
       \AND 
       \name Anthony Quinn \email a.quinn@imperial.ac.uk \\
       \addr Imperial College London and Trinity College Dublin
       \AND 
       \name Fadwa Idlahcen \email idlahfad@fel.cvut.cz \\
       \addr Czech Technical University in Prague
       \AND 
       \name Yiran Chen \email       yiran.chen@duke.edu \\ 
       \addr Duke University}
\editor{My editor}

\maketitle

\begin{abstract}%

Dataset distillation (DD) is an increasingly important technique that focuses on constructing a synthetic dataset capable of capturing the core information in training data to achieve comparable performance in models trained on the latter. While DD has a wide range of applications, the theory supporting it is less well evolved. New methods of DD are compared on a common set of benchmarks, rather than oriented towards any particular learning task. In this work, we present a formal model of DD, arguing that a precise characterization of the underlying optimization problem must specify the inference task associated with the application of interest. Without this task-specific focus, the DD problem is under-specified, and the selection of a DD algorithm for a particular task is merely heuristic. Our formalization reveals novel applications of DD across different modeling environments. We  analyze existing DD methods through this broader lens, highlighting their strengths and limitations in terms of accuracy and faithfulness to optimal DD operation.  Finally, we present numerical results for two case studies important in contemporary settings. Firstly, we  address a critical challenge in medical data analysis: merging the knowledge from different datasets composed of intersecting---but not identical---sets of features, in order to construct a larger dataset in what is usually a small sample setting. Secondly, we consider out-of-distribution error across boundary conditions for physics-informed neural networks (PINNs), showing the potential for DD to provide more physically faithful data. By establishing this general formulation of DD, we aim to establish a new research paradigm by which DD can be understood and from which new DD techniques can arise.
\end{abstract}


\date{December 2023}

\section{Introduction} \label{sec:intro}

 Dataset distillation (DD) refers to a range of techniques for generating a synthetic dataset whose sample size $N_S$ is relatively small, that functionally represents a much larger ($N_T\gg N_S$) training set. The functional requirement is that  a machine learning model trained on the synthetic data perform comparably in a specific task to one trained on the original data. DD has attracted increasing interest in recent years~\cite{wang2018dataset,zhao2020dataset}, with the growing research activity  reflected in several survey articles~\cite{lei2023comprehensive,yu2023dataset,geng2023survey}.

Various methods have been developed for DD. Based on how the synthetic data is optimized, these methods can be broadly categorized into two approaches: 
\begin{enumerate}
    \item designing new criteria to measure the similarity between synthetic and original datasets, \textit{e.g.},~\cite{zhao2020dataset,zhao2023dataset,kim2022dataset,cazenavette2022dataset,liu2023dream}; and 
    \item calculating meta validation loss on original datasets with models trained on synthetic ones, which can be regarded as the meta-learning style, \textit{e.g.},~\cite{wang2018dataset,nguyen2020dataset,nguyen2021dataset,loo2022efficient,zhou2022dataset}. 
 \end{enumerate}
 However, the current body of research lacks a cohesive framework. Formal proofs connecting the proposed criteria to a general notion of DD performance are missing, and there is a lack of mathematical and statistical insights in the existing literature. We argue that this gap exists because the task of DD is not yet well defined. Our goal is to address this issue by presenting a generalized framework that can provide a clearer conceptualization of the DD task.

Among the two primary approaches to DD, above, the first focuses on optimizing training efficiency, making it more practical and directly linked to performance outcomes. The optimization criteria within this approach can, in turn, be  categorized into two types: 
\begin{itemize}
    \item[(i)]
criteria which ensure distributional similarity between the synthetic dataset and the original training set, \textit{e.g.},~\cite{zhao2023dataset,zhao2023improved,sajedi2023datadam}; and  
\item[(ii)] criteria which target the training dynamics, specifically the gradients along the training trajectories, \textit{e.g.},~\cite{zhao2020dataset,cazenavette2022dataset,liu2023dream}. 
\end{itemize}
However, there has been little analysis to determine why certain criteria are more effective for DD than others, or how to establish robust benchmarks for evaluating whether a given criterion genuinely advances DD performance. Of course, there is clear intuition behind these methods, for distributional similarity suggests an independent sub-sample and training dynamics motivate some functional criteria through its gradient definition. Thus the methods' clear sensibility presents an opportunity for a proper evaluation as to the intuition and mechanism behind their merit.


One recent innovative but superficial formal model is provided in~\cite{sachdeva2023data}. Here, the evaluation criterion for DD is the test error, that is, the error on the population of a model trained on the synthetic dataset. Of course, this is never available, so DD methods can be considered various approximations to this ideal procedure. This perspective aligns with the bilevel optimization problem which has been adopted as a suitable framework for DD in other studies (e.g.~\cite{vahidian2024group} : \\
\begin{equation}\label{eq:bileveldd}
   \min\limits_{\hat{D}},\, f(\theta^*,D),\,\theta^* \in\arg\min f(\theta,\hat{D})
\end{equation}
where $f(\theta,D)$ is the loss of the model with parameters $\theta$ on the population data $D$ and $\hat{D}$ is optimal synthetic dataset. 

To evaluate the faithfulness of this criterion as an appropriate definition of the functional features and properties of DD, we consider a thought experiment, below. Specifically, we argue that it is possible to construct a solution that---despite satisfying criterion (\ref{eq:bileveldd})---is vacuous, and cannot possibly serve as a valid criterion for what can be commonly understood to be the intention of DD.


Recall that in standard DD works, the model is a Deep Neural Network (DNN), and, given that $N_S$ is small, is overparameterized relative to the distilled dataset, which implies that there exist multiple minimizers with zero loss for the inner optimization problem in (\ref{eq:bileveldd})~\cite{cooper2021global}. Consequently, the optimal dataset is one where
\[
\arg\min\limits_{\theta} f(\theta,\hat{D})\subset \arg\min\limits_{\theta} f(\theta,D)
\]
This inclusion suggests a vacuous solution, because it implies the following unreasonable anomalies:
\begin{enumerate}
    \item There is no guarantee that $\hat{D}\in \text{supp}(D)$.  Under some models, it may be very far from the support of the original population distribution $D=(X,Y)$ and still satisfy (\ref{eq:bileveldd});

    \item It is evident that increasing the size of $\hat{D}$ makes the problem more challenging, as the minimal surface becomes lower-dimensional with a larger sample size~\cite{cooper2021global}. Intuitively, however, a larger synthetic dataset allows for better representation of the distribution.
    
    \item The sufficient statistics of $D$ are enough to yield a solution. For instance, if $\hat{D}=\{x,y\}$ with $x=\mathbb{E}[X],\,y=\mathbb{E}[f(\theta^*, X)]$ with $\theta^*\in\arg\min\limits_{\theta} f(\theta,D)$ --this would be the optimal solution. 
\end{enumerate}

These anomalies suggest that the formulation of DD in (\ref{eq:bileveldd}) does not capture the full complexity of the uncertainty involved and the subtlety associated with DD practice and how it is done by ML practitioners. 



In this paper, we provide a formal definition of DD, which overcomes weaknesses in current approaches to the problem. A key insight is that criteria for DD should be tailored to the specific application of interest, rather than being predetermined as a one-size-fits-all approach. 

In summary, the main contributions of this paper are:
\begin{itemize}
    \item A formalization of the DD problem as on explicit optimization problem that is general and flexible as far as settings, criteria, models, data distribution and application.
    \item A comprehensive list of existing applications that can be subsumed in this framework, as well as new possible applications of this powerful technique.
    \item An understanding of the theoretical properties of the optimization problem and how existing techniques can be understood as approximating different components of the general procedure.
    \item Two case studies of novel applications in DD for settings of contemporary interest: medical data augmentation and physics informed machine learning generalization across boundary conditions. 
\end{itemize}

The rest of the paper is structured as follows: Section~\ref{sec:def} presents our formalism for the general DD optimization problem. In Section~\ref{s:ex}, we provide several examples, illustrating how this formalism provides intuitive flexibility to consider a range of applications of DD which have not yet been addressed in the literature.  Section~\ref{s:var} discusses variational aspects of the DD optimization problem, in light of the use of probability distributions as state variables. Section~\ref{s:classical} analyzes existing methods of DD in light of our  general framework. Section~\ref{s:cat} gives an analytical derivation of a DD solution, under our optimization framework, for categorical data, suggesting the possibility of its further development. In Section~\ref{s:dbnmed}, we present a novel and important application of DD: the synthesis of partially censored data in order to build a complete dataset appropriate for probabilistic graphical modelling in health applications. Another novel example is given in Section~\ref{s:phys}, involving physics-informed neural networks (PINNs). Finally, we provide concluding remarks in Section~\ref{s:conc}.

\section{The Formal Dataset Distillation (DD) Optimization Problem}\label{sec:def}

We now define the notation for all required quantities and operators:
\begin{itemize}
    \item $D=(D_X,D_Y)=(X,Y)$ is a random variable which represents sampling from the population from which the training set is drawn, associated with a probability space, $(\Xi,\mathcal{F},\mathbb{P}_D)$. For instance, in computer vision, $X$ can be pixel data and $Y$ the label  on an image. We can consider that $\mathbb{P}_D$ has a density $\mu$, with respect to an appropriate reference measure, $\lambda$ (typically Lebesgue or counting), $\mathbb{P}_D(A)=\int_{A} \mu(x,y) d\lambda(x,y)$ for $A\subset\mathbb{R}^{d_x+d_y}$. The domain of $(X,Y)$ can also be a manifold, $\mathcal{M}$. Also, in the generative case, the probability model may only involve the features $D=X$. For simplicity of notation, in the sequel we consider that $\Xi\subset\mathbb{R}^n$ and that every independent sample is in this finite dimensional Euclidean space.  
    \item A training set, defined as an i.i.d. sample of the population $\mathcal{T}=\{(x_i,y_i)\}_{i=1}^T\subset D$.
    \item We denote  the synthetic dataset by $\hat{D} = \{(\hat{x}_i,\hat{y}_i)\}_{i=1,...,N_S}$. This  finite-dimensional quantity  is the primary optimization decision variable in DD. 
    \item In the general case, consider $\mathcal{D}_{\theta(\omega)}(\cdot\vert\vert\cdot)$ to be the kernel of a probability distance or f-divergence, and thus takes two measures as arguments. In the sense of kernel we mean a distance $d$ on metric spaces, that is with $\mathcal{I}(D,D)=\mu^I$, a probability measure for the inference $I$, $\mathcal{D}_{\theta(\omega)}(\hat{I}\vert \vert \mathcal{I}(D,D))=\mathbb{E}_{\omega}l(y_1(\nu),y_2(\nu))=(y_1(\nu)-y_2(\nu))^2\in \mathbb{R}^+\times\Xi$. That is, for a random element in $D$, we define the metric distance from the ideal to the computed distribution of the inference task. When the latter is itself stochastic, we take the expectation. 
    \item $M_{\theta}\in \mathcal{B}$ is a model in some function space parameterized by $\theta$,  for instance a (deterministic) linear function, $y=\beta_0+\beta_1^T x$. Thus, formally, $M_{\theta}$ is a map from the parameters $\theta$ into a function space, ${\mathcal B}$, \textit{e.g.} $\left\{(\beta_0,\beta_1)\in\mathbb{R}^2\right\}\to \beta_0+\beta_1^T x$. Alternatively, $M_{\theta(\omega)}\in \mathcal{B}(\omega)$ is either the optimization problem with noise or error, or a Bayesian model. For instance, a neural network $\sigma(x, \theta(\omega))$ with weights following some parameterized distribution or process, $\theta(\omega)\sim \mathcal{W}_{\omega}(w)$ with hyperparameters $w$. Thus, formally, $M_{\theta(\omega)}$ is  a stochastic process defined on a function space. We shall sometimes use, e.g.,~ $M[\theta]$ instead as for notational convenience.
    \item $O(\cdot)$ or $O(\cdot,\omega)$ denotes an optimizer with guaranteed convergence to a minimum, or an ergodic sampler for Bayesian models. For instance, consider the application of stochastic gradient descent (SGD) on $\bar{D}\subset D$ yielding $O(\bar{D})\to \hat{\theta}$, where now $\hat{\theta}$ are the neural network weights to use in $M_{\theta}$, the neural network model. In the case of a sampler rather than a point optimizer, a canonical example is the  application of Hamilton Monte Carlo, yielding a Gibbs distribution on the same problem, i.e.\ $\theta\sim \omega = \frac{e^{-f(\bar{D})}}{Z}$, a posterior or pseudo-posterior. In this case, we incorporate the additional stochastic parameterization, $M_{\theta(\omega)}$, and so inference on input variables itself corresponds to sampling, that is, first from $\theta(\omega)$ and then application of this instance on the input to yield $M_{\theta(\omega)}[\bar{x}]\to \bar{y}(\omega)$
    \item $\mathcal{R}_{D_X}$ denotes a risk measure, which can be the risk-neutral expectation operator, $\mathbb{E}[\cdot]$, or a specific  risk measure such as the conditional value at risk (CVaR). The distribution with respect to which the expectation oeprator in the risk quantity is taken is the marginal $\mu_x$, i.e.\ the distribution of the input variable, $X$.
    \item $\mathcal{I}(M_{\theta},D)$ is an inference task of the model on the population data $D$. The output yields a particular distribution, possibly degenerate (i.e.\ delta-type).   An example is a collection of evaluations of a trained model on a set of input data. Another example is a condition or marginal inquiry with respect to some of the input variables of the posterior likelihood, i.e.\ the distribution of $Y$. $\mathcal{I}(\cdot,\cdot)$ could also be itself the input to an optimization or sampling problem, such as in neural architecture search (NAS). Formally, $\mathcal{I}:\mathcal{B}\times \Xi \to \bar{\mu}^I_D$, where $\bar{\mu}^I_D$ denotes a measure on $\Xi$  With an abuse of notation, we also write $\mathcal{I}(D,D)$ as some conditional marginalization of the distribution of $D$, but generically we write $\mathcal{I}(g,D)$ where $g:\Xi\to \Xi$ is some operator on $D$. \end{itemize}
With these notions, we  define dataset distillation (DD) as the following optimization problem:
\begin{equation}\label{eq:ddopt}
    \begin{array}{rl}
    \min\limits_{\hat{D}\in\mathbb{R}^{S\times n} } & \mathcal{R}_{D_X}\left\{\mathcal{D}_{\theta(\omega)}\left(\mathcal{I}\left(M[O(\hat{D},\omega)],D \right) \vert\vert \mathcal{I}(g,D )\right)\right\}
    \end{array}
\end{equation}
With this compact notation, we see that the DD optimization problem is defined as one of seeking a dataset, $\hat{D}$, which---if  used to train a model via $O$---ensures that a {\em specified\/}  inference task, ${\mathcal I}(\cdot, \cdot)$, exhibits minimal discrepancy from the true population statistics, averaged over noise in the statistical model. While the  decision variable of DD is $\hat{D}$, a deterministic quantity, the composition involves a nested sequence of probabilistic operations. Moreover, the regularity of the operations, that is their appropriate functional space and continuity properties, are not immediately clear. For instance, $O(\cdot)$ may not  even be a continuous function of its argument, $\hat{D}$, for example if it involves minimization of a non-convex loss function. The criteria and algorthmic choices for $O$ can significantly affect the outcome of inference $\mathcal{I}$. This clearly suggests that the inference accuracy---or some surrogate thereof---should be of prime consideration in the choice of design of the operator $O$. Ultimately, though, because little is even qualitatively certain regarding the map from the dataset $\hat{D}$ to the quantity given as the full cost function, that is, $\mathcal{R}_{D_X}\{\mathcal{D}_{\theta(\omega)}(\cdots$ above,  in~\eqref{eq:ddopt}, there are significant challenges in demonstrating the existence of a minimizing sequence and a limit of that sequence. 

Let us now write the optimization problem lifted to include equality constraints. Formally: 
\begin{equation}\label{eq:ddoptlift}
    \begin{array}{rl}
    \min\limits_{\hat{D},\hat{\theta}(\omega),\hat{I}} & \mathcal{R}_{D_X}\left\{\mathcal{D}_{\omega}\left(\hat{I}(\omega) \vert\vert \mathcal{I}(g,D)\right)\right\}\\ 
    \text{subject to } & \hat{\theta}(\omega) = O(\hat{D},\omega) \\ 
    & \hat{I}(\omega) = \mathcal{I}(M_{\hat{\theta}(\omega)},D)
    \end{array}
\end{equation}
Thus, we interpret a dataset as properly distilled if it is useful for some inference task $\mathcal{I}$, that is the model trained on it will yield a marginal distribution $\hat{I}$ that closely approximates a desired functional on the population distribution $\mathcal{I}(g,D)$. As this could be applied in a variety of domains for a variety of purposes, we aim to analyze a general inference task. This permits practitioners to tailor optimizers to the specific inference task and application of interest. We present the equality constraints explicitly in (\ref{eq:ddoptlift}), rather than implicitly in the objective function (\ref{eq:ddopt}), in order to reveal the derivative structure which might be exploited in the  optimization. 

This decomposition presents two additional state variables, $\hat{\theta}(\omega)$ and $\hat{I}$. The first one presents a flexible set of possibilities:
\begin{enumerate}
    \item $\theta\in\mathbb{R}^m$, for a point estimate vector of an optimal set of parameters
    \item $\theta(\omega)\in \mathcal{M}(\mathbb{R}^m)$, indicating that it is a probability distribution over the $m$-dimensional real-valued parameter vector.
    \item $M_{\theta(\omega)}\in \mathcal{M}(\mathcal{B}(\mathbb{R}^d))$, i.e., the most general case of a nonparametric model, where $\mathcal{B}$ is an appropriately chosen function space. In this case we have a stochastic process on the data vector space $\mathbb{R}^d$. 
\end{enumerate}
The quantity $\hat{I}$ is the output of the application of the model to the population data as far as computing some inference task. An inference task can simply be the test error, i.e.\ comparing the push-forward of the model on the input data, $M_{\theta(\omega)}(X)$, to the conditional distribution $Y\vert X$, weighed by the population prior of $X$. Otherwise, it can be some marginal or conditional enquiry, a likelihood or scoring function calculation, etc. 

Let us consider the parametric (possibly Bayesian) case first. Generically, both $\theta(\omega)$ and $\hat{I}$ are probability density functions, which are generally considered in the functional class of distributions, and so we must apply the appropriate analytic machinery to consider solution existence and regularity.

Note that the training set, $\mathcal{T}$, does not explicitly appear in either formulation of DD above (i.e.\ in (\ref{eq:ddopt}) or (\ref{eq:ddoptlift})). Rather, it remains implicit and defines a data generating procedure of $\mathcal{T}\sim D$ in order to evaluate the quantity $\mathcal{I}(g,D)$ in the objective. In other words, $\mathcal T$ is  a subset to be used for empirical risk minimization, or samples taken with or without replacement to mimic an online learning stochastic approximation setting. 

Formally, we  consider~\eqref{eq:ddoptlift} to be the \textbf{Online Stochastic Approximation} problem, i.e., we have unlimited access to i.i.d. samples of $D$. Then an 
\textbf{Empirical Risk DD} variant, with  limited training data, $\mathcal{T}$, becomes a sample average approximation (SAA) to operations within the problem; i.e. with $N_T$ training samples and $N_W$ samples of $\omega$: 
\begin{equation}\label{eq:empdd}
\begin{array}{l}
I^*=\mathcal{I}(g,D)\approx \mathcal{I}(g,\mathcal{T}), \\
\mathcal{R}_{D_X} [\mathcal{D}(\hat{I}\vert\vert I^*)] =\mathbb{E}_{D_X} [\mathcal{D}(\hat{I}(\omega)\vert\vert I^*)]=\approx \frac{1}{N_T N_W} \sum\limits_{j=1}^{N_W}\sum\limits_{i=1}^{N_T} d(\hat{I}(\omega_j),g(x_i,y_i))
\end{array}
\end{equation}
While it may then be tempting to redefine~\eqref{eq:ddoptlift} in terms of the sample averages, we avoid this in light of the fact that sub-sampling may be useful as intermediate procedures to  target the real risk more successfully (see, \textit{e.g.}~\cite{patil2024generalized} and, in application to DD directly,~\cite{vahidian2024group}).

We present several examples of the general DD formulation (\ref{eq:ddoptlift}) in the next Section. We include  some of the existing DD use cases, as well as some novel potential applications that this flexible formulation facilitates. 
Then, in Section~\ref{s:var}, we analyze the optimality conditions of the problem. The fact that we decompose the DD problem explicitly with respect to the constraints (\ref{eq:ddoptlift})---thereby introducing the measure-valued intermediate states explicitly in the problem---reveals a more transparent understanding of the solution, as well as an understanding of various potential approximations.

\section{Inference Tasks as DD applications}\label{s:ex}
In order to provide some grounding for the DD formalism presented in Section~\ref{sec:def}, we now consider several DD applications, and construct concrete examples of all of the operators and quantities given in~\eqref{eq:ddopt}.

\begin{example}
    \textbf{Test Error} This is the simplest setting, being the one invoked in most of the published numerical experiments  on DD. Furthermore, it is equivalent to the axiomatic DD optimization problem defined in~\cite{sachdeva2023data}.
    
    For simplicity, we assume an optimization procedure with global convergence guarantees, and so
    \[
    O(\hat{D}) = \arg\min\limits_{\theta} ,\, l(m(\theta,\hat{D}_x),\hat{D}_y)
    \]
    where $l$ is the loss function and $m$ is the model associated to $O$. This same function is the target of inference:
    \[
    \left(\mathcal{I}:X\to Y\right)[D_x],\,\hat{I}=m(\theta,D_x)
    \]
    If we take $\mathcal{R}=\mathbb{E}$  and $\mathcal{D}=l(\cdot,\cdot)$ in (\ref{eq:ddopt}), we finally get the problem:
\begin{equation}\label{eq:ddoptex1}
    \begin{array}{rl}
    \min\limits_{\hat{D},\hat{I},\theta} & \mathbb{E}_{X} \left[l(\hat{I},D_Y)\right]\\ 
    \text{subject to } & \theta = \arg\min l(m(\theta,\hat{D}_X),\hat{D}_Y) \\ 
    & \hat{I} = m(\theta,D_X)
    \end{array}
\end{equation}
As noted in Section~\ref{sec:intro}, we can construct a vacuous solution as follows: If we know some $\theta^* = \arg\min l(m(\theta,D_X),D_Y)$, and $\theta\in\mathbb{R}^d$, then with $\sharp(\hat{D})=N_S\ll d$, there is a $d-N_S$ dimensional manifold that satisfies $\theta^* = \arg\min l(m(\theta,\hat{D}_X),\hat{D}_Y)$. Any such $\hat{D}$ will solve~\eqref{eq:ddoptex1}. This suggests that the problem is insufficiently determined, and a more precise specification of the DD criteria will enable a more precise understanding of potential algorithmic solutions. 
\end{example}

\begin{example}
    \textbf{Conditional Marginal Uncertainty Quantification} This is   similar  to the previous example, but points to the flexibility of this DD construction. It can be motivated by considerations in medical diagnosis. We want to estimate the probability of a positive label under the condition that one input variable is a particular value. Furthermore, we would like to consider the entire uncertainty profile of the label across different inputs. We use an HMC sampler with a stationary distribution $\pi^{\theta}(V)$ of Gibbs form on the potential $V$. In this case, formally,
    \begin{equation}\label{eq:ddoptex2}
    \begin{array}{rl}
    \min\limits_{\hat{D},\hat{I},\theta} & \mathbb{E}_{X} \left[\mathbb{E}_{\omega}\mathcal{KL}(\hat{I}(\omega),\mathcal{I}(\{y\vert D_X :[X]_i=a\},D_Y)\vert D_X :[X]_i=a\right]\\ 
    \text{subject to } & \theta \sim \pi (m(\theta,\hat{D}_{X\vert [X]_i=a}),\hat{D}_Y,\omega) \\ 
    & \hat{I}(\omega) = \pi^{\theta} (X \vert [X]_i=a,\omega)
    \end{array}
\end{equation}
where here we abuse notation and simply mean the $\mathcal{KL}$ kernel, that is, $\mathcal{KL}(a,b)= a\log(b/a) $
\end{example}

\begin{example}
    \textbf{Neural Architecture Search} In this case we are interested in a dataset that is useful insofar as optimizing an inner problem in a bilevel optimization problem. The inner problem considers the optimal weights given an architecture. In this case the inference depends on the solution of an optimization problem (the inner problem). Consider a prior $\pi^{\mathcal{A}}$ over the architectures before their optimization, we want the dataset that fits the best performing weights across possible architectures $\mathcal{A}$ defining $m$, i.e., it is itself a measure on a function space $\mathcal{B}$. This can be constructed as follows:
\begin{equation}\label{eq:ddoptex3}
    \begin{array}{rl}
    \min\limits_{\hat{D},\hat{I},\theta} & \mathbb{E}_{X} \left[\mathbb{E}_{\mathcal{A}}[l(\hat{I}(\mathcal{A}),D_Y)]\right]\\ 
    \text{subject to } & \hat{\theta}(\mathcal{A}) = \arg\min l(m(\theta,\hat{D}_X,
    \mathcal{A}),\hat{D}_Y) \\ 
    & \hat{I}(\mathcal{A}) \sim \pi^{\mathcal{A}}\circ \left[\hat{\theta}(\mathcal{A}),D_X\right]
    \end{array}
\end{equation}
where the last quantity is the distribution, with respect to the prior on $\mathcal{A}$, of the predicted label given $\theta$ and an element of the distribution of $X$.
\end{example}
\begin{example}
\textbf{Continuous Learning} Consider that we want to be able to update a model to future heterogeneous data, while simultaneously continuing to be able to fit historical data, i.e., avoid catastrophic forgetting. We can formulate the procedure by indicating that the optimizer will be parameterized to add an additional dataset with a shifted distribution. 
\begin{equation}\label{eq:ddoptex4}
    \begin{array}{rl}
    \min\limits_{\hat{D},\hat{I},\theta(\omega)} & \mathcal{R}_{D_X}\left\{\max\left(\mathcal{D}_{\omega}\left(\hat{I}(\omega) \vert\vert \mathcal{I}(\bar{D}(\omega),\cdot )\right),\mathcal{D}_{\omega}\left(\hat{I}(\omega) \vert\vert \mathcal{I}(D,\cdot )\right)\right)\right\}\\ 
    \text{subject to } & \hat{\theta}(\omega) = O\left(\{\hat{D}\cup\bar{D}(\omega)\}\right) \\ 
    & \hat{I}(\omega) = \mathcal{I}(M_{\hat{\theta}(\omega)},D)
    \end{array}
\end{equation}
The generic inference operation $\mathcal{I}$ is flexible, but can be (test) loss. The subscript of $\mathcal{D}$ over $\omega$ indicates the distribution over past data considered, and thus the expectation performs a weighing of the historical loss. 
\end{example}
With this framework, one can obtain the appropriate intuition as to how to design DD algorithms for a given task. While all of these examples are intractable as they are stated, each component can be approximated using well known stochastic optimization techniques associated with Stochastic Approximation and Sample Average Approximation. In this way the training samples are used $\mathcal{T}$ algorithmically to approximate $D$. Thus algorithms can be developed to increasingly improve the various aspects of the training over time, escalating the research program of DD.

We present three final examples in greater detail. The first two we highlight as the target of this work and the focus of the Numerical Results. The final example includes a comprehensive analytic derivation of the details for pedagogical purposes.
\begin{example}
\textbf{Medical Data Bootstrapping} It often occurs in applications involving public health and modeling medical diagnosis and treatment that a number of heterogeneous data sets are available. At the same time each particular dataset is one with a relatively low number of individual samples, due to the cost of scaling the participant count in clinical trials, that is relatively high dimensional, as medical advances has yielded a number of informative biomarkers one can measure. As far as modeling from the ML perspective, the issue of overparameterization becomes paramount, and so we propose an approach that \emph{leverages} the large union of many different data sources to construct an accurate neural model, then constructs a dataset that is informative for the express purpose of constructing a probabilistic graphical model (PGM),
\begin{equation}\label{eq:ddoptex4}
    \begin{array}{rl}
    \min\limits_{\hat{D},\hat{I},\theta(\omega)} & \mathcal{R}_{D_X}\bigcup\limits_{i}\left\{\mathcal{D}_{\omega}\left(\hat{I} \vert\vert \mathcal{I}(D,D(\tilde{y}_i))\right)\right\}\\ 
    \text{subject to } & \hat{\theta}(\omega) = O_{PGM(i)}\left(\hat{D},\omega\right) \\ 
    & \hat{I} = \mathcal{I}(M_{\hat{\theta}(\omega)},D(\iota_i))
    \end{array}
\end{equation}
where $\{\iota_i\}$ is a set of possible inputs and $\{\tilde{y}_i\}$ the outputs of the tasks of inference of interest, \textit{e.g.}, expected time to disease progression, conditional treatment effect, etc. with the associated marginal distribution $D(\iota_i)$, $O_{PGM}$ performs structure and parameter learning from the data-set $\hat{D}$. We let this depend on the particular inference task, as different structures of PGMs are more efficient to perform different tasks. We consider generically a risk over the union of such inference measures, and in practice this can be some weighted average across the choices of inference tasks.
Note that this formulation does not explicitly incorporate the neural modeling, this will be described as an instrumental technique. 
\end{example}

\begin{example}
\textbf{Physics Informed Machine Learning} 
Consider that we have some PDE $\mathcal{F}(u(x),\gamma(x))=0$ on $\Omega$ with boundary conditions $\mathcal{B}(u(x))=0$ for $x\in\partial\Omega$. A neural network is used to parameterize $u$ to obtain a loss function, over a dataset of $N$ lattices $\{x_k\}\in\mathbb{R}^{d\times M}$ and their corresponding values $\hat{f}_k$
\[
\mathcal{L}_F(\theta_u,\theta_{\gamma};\{x\},\hat{f})=\frac{1}{MN}
\sum\limits_{m=1}^M\sum\limits_{l=1}^N (\mathcal{F}(\bar{u}_l(\theta,x_m),\bar{\gamma}_l(\theta,x_m))-\hat{f}_{ml})^2
\]
Suppose that we are particularly interested in generalization across boundary conditions. As such, consider that there is a population prior on the boundary $\pi\left(u_{\partial\Omega}\right)$. Based on this prior we would like to minimize the chance that a set of parameters $\hat{\theta}$ trained on $\hat{D}$ minimizes the Conditional Value at Risk, across the distribution , of the PDE interior reconstruction loss. Consider that $\hat{D}_g$ also includes a set of samples of functions $g(x)$ with $x\in\partial\Omega$. We can assume that $g$ is sampled from $\pi\left(u_{\partial\Omega}\right)$,

\begin{equation}\label{eq:ddoptex6}
    \begin{array}{rl}
    \min\limits_{\hat{D},\theta} & \mathbf{CVaR}_{\pi\left(u_{\partial\Omega}\right)} \left[\mathcal{L}_F(\theta_u,\theta_{\gamma};D)\right]\\ 
    \text{subject to } & \theta = \arg\min \mathcal{L}_F(\theta_u,\theta_{\gamma};\hat{D}) \\ 
    & \qquad \text{s.t. } u(x_{\partial\Omega},\theta,g) = g(x_{\partial\Omega}) \sim \hat{D}_g 
    \end{array}
\end{equation}
where we can collapse the inference task $\hat{I}$ as just the application of the neural network on the distributional data (which we would use the training data to estimate).
\end{example}

Finally, we present a more transparent analytic derivation of the learning optimization problem defined by DD towards a classical approach to modeling time series data. 

\begin{example}
    \textbf{Mixture of AR Time Series Models} Inspired by~\cite{karny2016recursive,smidl2005mixture}. As a reasonable concrete setting in which DD can apply, consider that we wish to learn a mixture of Autoregressive Time Series models for a multi-dimensional stochastic process, defined over mixture weights $\{\mathbf{m}_q\}$ and transition matrices $\{\mathbf{M}_q\}$,
    \[
    X_{k+1} \sim \sum\limits_{q=1}^m \mathbf{m}_q \mathbf{M}_q^T X_{k+1}
    \]
    where for simplicity we consider a finite mixture. In the case that $X_k$ is categorical, that is it is $\mathcal{X}$-valued, where $\mathcal{X}$ is isomorphic to $\mathbb{Z}^+_{/n}$ for some $n$, this permits sufficient expressivity for any potential transition. 
    
    Consider that at a particular system is observed over a finite period of time. Then, the observations cease, either due to their being unobservable for a subsequent period of time, or the research staff being unavailable to collect the data, or even some technical malfunction. Knowing that the process is stationary, the learner constructs a synthetic dataset that is representative of the historical information. We consider that the desired inference task is forecasting the subsequent time period. 
\begin{equation}\label{eq:ddoptex7}
    \begin{array}{rl}
    \min\limits_{\hat{D},\hat{I},\theta(\omega)} & \mathcal{R}_{D_t}\left\{\mathcal{D}\left(\hat{I} \vert\vert D_{t+1}\right)\right\}\\ 
    \text{subject to } &\mathbf{m} = O_{EAR}\left(\hat{D}\right),\,M\in \Delta^I \\ 
    & \hat{I} = \mathcal{I}\left(\sum \mathbf{m}_q \mathbf{M}_q,D_{t}\right)
    \end{array}
\end{equation}
which we solve with a standard Empirical Risk least squares Minimization as,
\begin{equation}\label{eq:ddoptex7emp}
    \begin{array}{rl}
    \min\limits_{\hat{D},\hat{I},\theta(\omega)} & \left\|\hat{I} - \mathcal{T}_{t+1}\right\|^2\\ 
    \text{subject to } & \mathbf{m} = O_{EAR}\left(\hat{D}\right),\,M\in \Delta^I \\ 
    & \hat{I} = \mathcal{I}\left(\sum \mathbf{m}_q \mathbf{M}_q,\mathcal{T}_{t}\right)
    \end{array}
\end{equation}
We can obtain a consistent estimator $O_{EAR}$ with counting operations on the transitions. This assigns a model $\sum\limits_{i}^I m_i(\hat{D}) \Theta_i$ of mixture weights. From this model we obtain the forecast on the training data $\hat{I}=\sum\limits_{i}^I m_i(\hat{D}) \Theta_i^T X_{\mathcal{T}_t}$ and compute the reconstruction loss to the real future in the training $X_{\mathcal{T}_{t+1}}$. 
This is one function evaluation. However, unlike the other examples, there is no clear backwards pass or gradient, indeed $\hat{D}$ is categorical, then there is no well defined gradient.

In the subsequent Section~\ref{s:cat} we approach the problem of DD for categorical variates using integer programming. 
    
\end{example}

\section{Properties of the Optimization Problem}\label{s:var}
As noted in Section~\ref{sec:def}, despite the technical challenges involved, understanding the mechanics of DD requires studying a lifted variant of the problem wherein the intermediate operations are clearly visible and take on auxiliary variables. This decomposition allows a more fine tuned algorithm design, that is by focusing on particular components of the composition. Many specific ML techniques can be defined as approximations to some of these compositions. However, there is a significant difficulty added, namely that this becomes
infinite dimensional optimization problem on the set of probability measures.

In this section we derive and study the optimality conditions for the DD optimization problem defined above. Treatment of statistical fitting with distributions from an optimization context is classical in Bayesian analysis~\cite{csiszar1975divergence}. But
with $\hat{D}$ a finite dimensional real vector, in general variational considerations are well amenable to establishing optimality results for standard Bayesian inference techniques for parametric models, as the problem can be formulated simply as a function of the parameters in the model as decision variables~\cite{bissiri2016general}. When (even auxiliary) variables are introduced that are probability densities themselves, the machinery has to properly account for them. 

The study of optimization problems with distributions as explicit variables has been a growing topic of importance. The framework of probabilistic design~\cite{karny2012axiomatisation} explicitly seeks distributions to minimize distance to some target, under some conditions. Recently,  generalized variational inference~\cite{knoblauch2022optimization} codified a general approach to probabilistic modeling that encapsulates standard sampling, classical Variational Inference techniques, and others. 

Deeper formalisms include recent work considering the circumstance in which the measure space is endowed with the Wasserstein metric, thus presenting a topological vector space for which optimality conditions can be valid. See, \textit{e.g.}~\cite{molchanov2004optimisation} and a more recent presentation incorporating constraints~\cite{bonnet2019pontryagin}. More general techniques for incorporating a variety of functionals on distributions can be performed with Orlicz spaces~\cite{edgar1992stopping}.

Optimality conditions are generally conditions involving the directional derivative of the objective along feasible directions. As we are not considering any inequality constraints, feasible directions are just linearized subspaces of the equality constraints, that is tangent cones are not a necessarily technical focus point of our analysis. Extensions, however, are available, the appropriate variational analysis is described in~\cite{molchanov2000tangent}. 

Let us recall the formal optimization problem for reference, simplifying the presentation by considering point optimization rather than Bayesian models, i.e. $\theta(\omega)=\theta$, we obtain:
\[
    \begin{array}{rl}
    \min\limits_{\hat{D},\hat{I},\theta(\omega)} & \mathcal{R}_{D_X}\left\{d\left(\hat{I} , \mathcal{I}(D,\cdot )\right)\right\}\\ 
    \text{subject to } & M_{\theta(\omega)} = O(\hat{D}) \\ 
    & \hat{I} = \mathcal{I}(M_{\theta(\omega)},D)
    \end{array}
\]
where $d$ is simply a metric on the topological vector space corresponding to $\hat{I}$. 

We proceed with the analytical derivation of the directional derivatives. To this end the tangent cone corresponding to feasible directions $(\delta \hat{D},\delta\hat{I},\delta\theta)$ are the linearization of the constraints, that is,
\begin{equation}\label{eq:tancone}
    D M_{\theta^*}[\delta \theta] = D O(\hat{D})[\delta \hat{D}],\qquad
    \delta \hat{I}  =D^{(1)}\mathcal{I} \left(M_{\theta^*},D\right) D M_{\theta^*}[\delta \theta] 
\end{equation}
where the variations can be considered as Gateaux derivatives. The first expression indicates the linear sensitivity of the optimization solution $\theta^*$ in the direction of $\hat{D}$ (see~\cite{bonnans2013perturbation} for a comprehensive study, or~\cite{levy2001solution} for a concise informative presentation). That is, as the data parameterizing the optimization solver $\hat{D}$ is perturbed by a small variation, this induces a change in the solution $\delta\theta$ that perturbs the model $M_{\theta}$. We can observe that with most simple functional models, the parametric dependence on $\theta$ is Lipschitz or their gradient is Lipschitz. Moreover,  by adding a small strongly convex regularization to the objective, a unique (local) solution as well as local stability and directional differentiability subject to perturbations is expected to hold. 

The second expression involves the variation in the inference distribution with respect to the model. The overall stability will again depend on the specific inference computation. However, standard conditional, marginal, or total expectation operators are stable to first order parametric perturbations~\cite{dupavcova1990stability}. 

Continuing to variations of the objective, we obtain the stationarity condition:
\begin{equation}\label{eq:optcondsa}
    \partial\mathcal{R}_{D_X}\left\{d\left(\hat{I}^* , \mathcal{I}(D,\cdot )\right)\right\} D^{(1)}d\left(\hat{I}, \mathcal{I}(D,D)\right)\delta \hat{I}\ge 0
\end{equation}
for all $(\delta \hat{D},\delta\hat{I},\delta\theta)$ satisfying~\eqref{eq:tancone}. Note that $\mathcal{R}$ is convex, so the subgradient is the conventional one. From the Orlicz perspective (see~\cite{javeed2023risk} for a contemporary application inspiring this analysis) we can consider that $\hat{I}\in L^1(\mathbb{R}^{d_x+d_y})$ and the dual operator $ D^{(1)}d\left(\hat{I}, \mathcal{I}(D,D)\right)\in L^s(\mathbb{R}^{d_x+d_y})$ for $s\in[1,\infty)$ if the (topological vector space) pairing is endowed with the weak topology and $s= +\infty$ with a weak$^*$ topology. Note that application to $\hat{I}$ requires $M_{\theta^*}\in (L^1(\mathbb{R}^{d_x+d_y})^*$ and we shall consider this in the sequel. Finally $\partial\mathcal{R}_{D_X}\left\{d\left(\hat{I}^* , \mathcal{I}(D,\cdot )\right)\right\}\langle\cdot,\cdot\rangle$ is a bilinear form. We can see by similar reasoning as by~\cite{javeed2023risk} and~\cite{ruszczynski2006optimization} that we can ensure, for well behaved metrics $d$ and inference tasks $\mathcal{I}$, that any local minimizer satisfies the Variational Inequality and that there exists at least one tuple $(\delta \hat{D},\delta\hat{I},\delta\theta)$ satisfying~\eqref{eq:optcondsa}.

Now, consider the application of~\cite[Theorem 2.1]{molchanov2004optimisation} to define the first order necessary conditions as satisfied by $\hat{D}^*$, $\mu^{\hat{I}}_*$ and $\theta^*$ together with the two Lagrange multipliers $y^M\in C^1(\Theta,(L^1(\mathbb{R}^{d_x+d_y}))^{**})$ and $y^I\in (L^1(\mathbb{R}^{d_x+d_y}))^*$. The variational in Lagrange multiplier form is the condition that for \emph{all} valid tuples $(\delta \hat{D},\delta\hat{I},\delta\theta)$ (i.e., satisfying~\eqref{eq:tancone}), that are measureable and preserving nonnegativity of the operator, satisfy,
\begin{equation}\label{eq:optcondsvi}
\begin{array}{l}
        \partial\mathcal{R}_{D_X}\left\{d\left(\hat{I}^* , \mathcal{I}(D,\cdot )\right)\right\} D^{(1)}d\left(\hat{I}, \mathcal{I}(D,D)\right)\delta \hat{I} -\left\langle y^I,\delta\hat{I}\right\rangle \ge  0\\
\left \langle y^M ,D M_{\theta^*}[\delta \theta] \right\rangle +\left\langle y^I,\mathcal{I}\left(D M_{\theta^*} [\delta \theta],D\right)\right\rangle \ge 0\\
    \left\langle y^M,D O(\hat{D})[\delta \hat{D}]\right  \rangle  \ge 0
\end{array}
\end{equation}
where we applied linearity of $\mathcal{R}$ with respect to its base measure.

We proceed along the lines of reasoning in~\cite{molchanov2004optimisation}.
Using the Radon-Nykodyn Theorem, we can consider the first order optimality conditions for the optimization problem in strong form, as in as equation systems rather than statements on directional derivatives. In this case we denote the measure associated with $\hat{I}$ as $\mu^{\hat{I}}$ with densities $p^{\hat{I}}=\frac{d\mu^{\hat{I}}}{d\lambda}$. 

\begin{equation}\label{eq:optcondsstrong}
\begin{array}{l}
        \partial\mathcal{R}_{D_X}\left\{d\left(p^{\hat{I}^*} , \mathcal{I}(D,D )\right)\right\} D^{(1)}d\left(p^{\hat{I}}, \mathcal{I}(D,D)\right)p^{\delta\hat{I}} -\left\langle y^I,p^{\delta\hat{I}}\right\rangle =  0\\
\left \langle y^M ,D M_{\theta^*}[\delta \theta] \right\rangle +\left\langle y^I,\mathcal{I}\left(D M_{\theta^*} [\delta \theta],D\right)\right\rangle = 0\\
    \left\langle y^M,D O(\hat{D})[\delta \hat{D}]\right  \rangle  = 0
\end{array}
\end{equation}
Additional complications arise with Bayesian methods for computing $\theta(\omega)$, which are beyond the scope of the current work. 

With these optimality conditions in place, we can now consider different algorithms for DD as targeting the satisfaction of these equations with varying levels of precision or approximation. 


\section{Classical Methods}\label{s:classical}
\subsection{DD by Trajectory Matching}
We highlight Trajectory Matching because it appeared to be the most appropriate for analyzing within our framework, and furthermore it is generally considered the state of the art as far as numerical performance, \textit{e.g.},~\cite{cazenavette2022dataset,du2023minimizing,liu2024att}.

Consider the problem of finding a trajectory matching dataset $\tilde{D}$ defined in terms of SDE dynamics. The neural network model is given generically by the loss with respect to the parameter and data as $f(\theta,D)$. 

One possibility is to study the optimization problem of matching the gradients of the flows:
\begin{equation}\label{eq:gradmatch}
\begin{array}{rl}
    \min\limits_{\tilde{D}} & \mathbb{E}\int_0^T \left\|\nabla f(\theta_t,D)-\nabla f(\hat{\theta}_t,\tilde{D})\right\|^2 dt \\
    \text{s.t. }& d\theta_t = -\nabla f(\theta,D) dt+dW_t,\\
    & d\hat{\theta}_t = - \nabla f(\theta,\tilde{D})dt
    \end{array}
\end{equation}
Note that this is an optimization not control problem, i.e., $\tilde{D}$ is time independent. Also $\hat{\theta}$ has deterministic dynamics since gradient descent can be performed with the full dataset on the smaller synthetic data.

Writing the optimality conditions, we see:
\[
\mathbb{E}\int_0^T \nabla^2_{\theta D} f(\hat{\theta}_t,\tilde{D})\left( \nabla f(\theta_t,D)-\nabla f(\hat{\theta}_t,\tilde D)\right)dt - \int_0^T \nabla^2_{\theta D} f(\hat{\theta}_t,\tilde{D}) \lambda_t dt =0
\]
from which follows, using Fubini's theorem,
\[
\begin{array}{l}
\int_0^T \nabla^2_{\theta D} f(\hat{\theta}_t,\tilde{D})\mathbb{E}_D\left( \nabla f(\theta_t,D)-\nabla f(\hat{\theta}_t,\tilde D)-\lambda_t\right)dt = 0  \\
\int_0^T \nabla^2_{\theta D} f(\hat{\theta}_t,\tilde{D})\left( \mathbb{E}[\nabla f(\theta_t,D)]-\nabla f(\hat{\theta}_t,\tilde D)-\lambda_t\right)dt = 0
\end{array}
\]

Alternatively we can consider not matching the gradients but the actual trajectory eschewing random operators entirely by working with the Fokker Planck equation defining the distributional dynamics of $\theta_t$, defined by $\rho(\theta,t)$. This can done with a cosine similarity functional of the form:
\begin{equation}\label{eq:trajmatch}
\begin{array}{rl}
    \min\limits_{\tilde{D}} & \int_0^T\int_{\theta} \sigma_{cos}\left(\frac{\partial \rho(\theta,t)}{\partial t},(\theta-\hat{\theta}_t)\odot \frac{d\hat{\theta}_t}{dt}\right) d\theta dt \\
    \text{s.t. }& \frac{\partial \rho(\theta,t)}{\partial t} = -\nabla\cdot(\nabla f(\theta_{\rho},D))+\Delta \rho(\theta,t),\\
    & \frac{d\hat{\theta}_t}{dt} = - \nabla f(\theta,\tilde{D})
    \end{array}
\end{equation}
With optimality conditions,
\[
\int_0^T\int_{\theta} \nabla^2_{\theta D} f(\hat{\theta}_t,\tilde{D})\odot (\theta-\hat{\theta}_t)\odot \sigma'_{cos,(2)}\left(\frac{\partial \rho(\theta,t)}{\partial t},-(\theta-\hat{\theta}_t)\odot \nabla f(\hat{\theta}_t,\tilde{D})\right) d\theta dt =0 
\]

We can specify $f$ to be more specifically, in the standard mean field training regime notation~\cite{mei2018mean},
\[
f(\theta,\rho_t)=V(\theta)+\int U(\theta,\theta')\rho_t(d\theta'),\,V(\theta)=-\mathbb{E}[y\sigma(x;\theta)],\,U(\theta,\theta')= \mathbb{E}[\sigma(x;\theta)\sigma(x;\theta')]
\]
where $\sigma$ is the neural network model. The results of~\cite{mei2018mean} can be combined with the level of accuracy achieved in reconstructing the trajectory to obtain similar trajectory bounds with respect optimal distributional dynamics. 

\paragraph{Point of View from Stability}
Another possible perspective can be derived from~\cite{ho2020instability}.

In this work they consider instead of the standard empirical risk minimization and generalization decomposition, with $G$ the iteration
\[
G^t_k(\theta^0)-\theta^*=G^t_k(\theta^0)-\hat{\theta}_{ERM}+\hat{\theta}_{ERM}-\theta^*
\]
one analyzes a version of the algorithm with full data information, and perturbations from that,
\[
G^t_N(\theta^0)-\theta^*=G^t(\theta^0)-\theta^*+G^t_{N}(\theta^0)-G^t(\theta^0)
\]
and proceed to study algorithms to classify them as \emph{fast} or \emph{slow} and \emph{stable} or \emph{unstable} based on the optimization convergence rate for the full data dynamics $G^t-\theta^*$ and the stable tracking of the empirical dynamics to the full data dynamics $G^t-G_N^t$. It appears that DD is directly aiming to approximate the second term, 

\paragraph{Relation To DD Optimization Problem}

Ultimately it is clear that we can consider this to be attempting to satisfy the condition:
\begin{equation}\label{eq:trajmatchO}
    O(\hat{D})=O(\bar{D})\approx O(D)
\end{equation}
where $\bar{D}$ is a large training set. We observe that if the optimization process $O$ is such that,
\[
O(D)\subseteq \arg\min_{\theta}\mathcal{R}\left\{\mathcal{D}\mathcal{I}(M_{\theta}[O(D)],D_X)\vert\vert \mathcal{I}(D,D)\right\} 
\]
and moreover, that in the overparametrized distributional regime~\cite{simsek2021geometry},
\[
\min_{\theta}\mathcal{R}\left\{\mathcal{D}\mathcal{I}(M_{\theta}[O(D)],D_X)\vert\vert \mathcal{I}(D,D)\right\} = 0
\]
then the choice of targeting~\eqref{eq:trajmatchO} is reasonable for the original problem.

The fact that we are performing this on a training set means that we can understand algorithmic trajectory matching, \textit{e.g.}, as ~\cite{cazenavette2022dataset}, as Sample Average Approximation (SAA) applied to~\eqref{eq:trajmatchO}. That is, the algorithms are discretizations of~\eqref{eq:trajmatch}. 

Notice that one can alternatively solve the stochastic optimal control problem directly. The accuracy of the solution to the distributional PDE control formulation~\eqref{eq:trajmatch} depends on the error of discretizing the optimality conditions of this system. The weak form suggests continuously differentiable finite elements can be used, and we can get an error rate of the order of $O(h^2)$, with $h$ the grid size.

\subsection{Other Methods}
\paragraph{DD by Distribution Matching}
Distribution matching for DD, introduced in~\cite{zhao2023dataset}, amounts to using some metric of distance to directly optimize $\hat{D}$ as proximity to $\mathcal{D}$, and in practice some finite training set $\mathcal{T}\subset\mathcal{D}$. This is performed by applying an encoder to generate latent code for the data samples and comparing the expected difference,
\begin{equation}\label{eq:distmatch}
\mathbb{E}\left\|\psi(\hat{z})-\psi(z)\right\|^2,\,\hat{z}\in\hat{D},,z\sim D
\end{equation}
Let us consider this strategy in light of the optimization problem~\eqref{eq:ddopt}. Notice that if $\hat{D}=D$, then the inference model would be one that is optimized for $D$, i.e., $\hat{I}=\mathcal{I}(O(D),D)$, and so blundly it would appear that this is a generically effective technique. Furthermore, the use of a latent encoder suggests that a model from which the encoded value is potentially interpretable or explainable in some way. 

However, this does not take into account the inference quality of interest. Given the small dataset size, we can expect the minimization of distance solution~\eqref{eq:distmatch} to have many minima. Knowledge as to the intention of the DD can properly refine the solution to be appropriate for that task. As far as a versatile strategy wherein we can see the error in the estimate as:
\[
\mathcal{I}(O(D),D)-\mathcal{I}(O(\hat{D}),D) \approx \mathcal{I}(O(D),D)+D^{1}\mathcal{I}(O(D),D)[O'(D)](\hat{D}-D)
\]
that is, with an error associated to the data the optimizer works with by the optimmizers' and inference operator's sensitivity.

\paragraph{Factorization Methods}
Factorization methods~\cite{liu2022dataset} are a recent popular competitive method, without much theoretical intuition provided. They consider the synthetic dataset to be simply a factorization wherein the original data is represented in a lower dimensional basis. This makes sense from the Neural Tangent Kernel (NTK) point of view~\cite{jacot2018neural}. That is, with $\hat{D}=b\hat{c}$ with $\|b\hat{c}-D\|$ minimized in some sense, then we can expect an optimizer will effectively project the gradient flow in the space of $b$. If the basis is chosen suitably to target $\mathcal{I}$, can result in the desired training performance.

\paragraph{Recent Work and Methodological Insights}
Finally, here we shall review recent variations of existing methods and state of the art work as of writing, and attempt to provide brief insight into these methods through our framework.

As stated previously, mainstream dataset distillation methods can be roughly categorized into matching-based ones and bi-level optimization ones, \textit{e.g.},~\cite{zhao2020dataset,zhao2023dataset,cazenavette2022dataset,nguyen2020dataset,loo2022efficient}. For matching-based methods, recent explorations mainly focus on designing better criterion, such that the synthetic data can better reflect the characteristics of original data.~\cite{zhang2024m3d},~\cite{shang2023mim4dd} and~\cite{deng2024iid} adopt Maximum Mean Discrepancy, mutual information and sample relationship to better fit the original distribution, respectively. These can be understood to establish proxy variables for the inference task of interest $\mathcal{I}$ into the otherwise completely agnostic distributional matching techniques, thus ameliorating the lack of steering problem described.

\cite{liu2024att} dynamically select trajectories with proper length to optimize the synthetic data. There are also works aiming at selecting informative samples for matching, \textit{e.g.},~\cite{liu2023dream,xu2024distill}. These can be readily understood to improve the accuracy of the approximation $O(\hat{D})\approx O(D)$. 

Regarding bi-level optimization methods, meta-learning is incorporated into the surrogate image update process, using validation performance as a direct optimization target to enhance the quality of distilled datasets. This approach has been explored extensively in recent works, where nested optimization loops are used to iteratively refine synthetic datasets while fine-tuning model parameters to achieve optimal performance on validation tasks, \textit{e.g.},~\cite{zhou2022dataset,loo2023dataset}. This can be understood to be ameliorating the vacuous solution problem identified with the bilevel approach at the beginning of our paper. By iteratively incorporating validation datasets, generalization is directly targeted, ensuring that at least for the purposes of representing data to optimize a model's population loss the technique is effective.

In addition to the above works, there are also methods designed from generative perspectives, \textit{e.g.},~\cite{cazenavette2023generalizing,gu2024efficient,su2024d4m}. Through introducing generative prior, the generated images demonstrates better cross-architecture generalization and interpretability. 
Efforts also include dataset parametrization, enabling more information in the restricted dataset size, \textit{e.g.},~\cite{kim2022dataset,wei2023sparse,liu2022dataset}. A generative prior corresponds to $\omega$ including some previously learned components to bias the model $M_{\hat(\omega)}$. Thus background knowledge can be seen to easily fit into the framework.

While dataset distillation achieves certain progresses in constructing small surrogate datasets, it is further applied to many other research topics, such as continual learning and federated learning, \textit{e.g.},~\cite{gu2024ssd, vahidian-dd-fl,yang2023efficient,xiong2023feddm,wang2024fed}.
The trustworthy issue is also broadly focused to make dataset distillation more practical in real-world scenarios.~\cite{vahidian2024group} involve a group distributionally robust optimization method to enhance the robustness of distilled data.~\cite{zhu2023calibration} mitigates the information loss during the distillation process, and better calibrate the data and finally~\cite{Lu_2024-bias-dd} investigates the influence of dataset bias on DD. These can be all understood as 1) acknowledging the fact that $\mathcal{T}$ is just a sample of $D$ and adopting measures of robustness to mitigate some of the practical consequences of sampling error and 2) adopting (potentially composite) criteria $\mathcal{I}$ to target particular secondary objectives, and 3) developing tools to improve the approximation $O(\hat{D})\approx O(D)$.

\section{Mixture Extended AutoRegressive Model and an Invitation of DD for Categorical Data}\label{s:cat}
Recall the Autoregressive Model given as the example at the end of Section~\ref{s:ex}
\[   
\begin{array}{rl}
    \min\limits_{\hat{D},\hat{I},\theta(\omega)} & \left\|\hat{I} - \mathcal{T}_{t+1}\right\|^2\\ 
    \text{subject to } & \mathbf{m} = O_{EAR}\left(\hat{D}\right),\,\mathbf{m}\in \Delta^m \\ 
    & \hat{I} = \mathcal{I}(\sum_Q \mathbf{m}_q\mathbf{M}_q,\mathcal{T}_{t})
    \end{array}
\]
We now describe how to compute $O_{EAR}$ with counting operations on the transitions to obtain a model $\sum\limits_{q}^m \mathbf{m}_q(\hat{D}) \mathbf{M}_{q}$ of mixture weights. 

We consider that there are 3 states $\mathbb{Z}/3\mathbb{Z}:=\{0,1,2\}$ for all $n$ features and $N$ data observations.
\[
\hat{D} = \{z_{i,l}\}\subset \left(\mathbb{Z}/3\mathbb{Z}\right)^{nN}
\]
Now consider that there is an unknown Markov transition matrix $\mathbf{M}$ that describes the evolution of $D$ and that this matrix is of the form:
\[
\mathbf{M}= \mathbf{m}\cdot \left[\mathbf{M}^{(1)},...,\mathbf{M}^{(m)}\right]
\]
i.e., it is a linear combination defined by a mixture $\mathbf{m}\in \Delta^m$ coefficient combination of the set of matrices $\{M_i\}$. Let $\bar{n}:=3^n$. Each matrix is valued $\bar{n}\times \bar{n}$ and is row stochastic. 

We initialize $\mathbf{m}$ to be a uniform Dirichlet prior, i.e., with
\[
\vec{\alpha}=\{\alpha\}^{m},\,\alpha = 1/m
\]
We obtain a prior:
\[
\pi(\mathbf{m}) = \frac{1}{\Gamma(\alpha)^{m}} \prod_{q=1}^{m} [\mathbf{m}]_q^{\alpha_{q}-1}
\]
We can make this a prior on the transition matrix itself for a given row $i$:
\[
\pi(\mathbf{M}_i) = \frac{1}{\Gamma(\alpha)^{m}}\prod\limits_{j=1}^{\bar{n}} \left\{\sum_{q=1}^{m} \left([\mathbf{m}]_q^{\alpha_{q}-1}\mathbf{M}^{(q)}_{ij}\right)\right\}
\]

This is conjugate to the prior for the categorical distribution. Thus upon obtaining $\hat{D}$ we can obtain a posterior estimate of the distribution. Consider some natural isomorphism $\phi$ (\textit{e.g.}, lexicographic) from $\{1,...,\bar{n}\}$ to the permutations of $\mathbf{Perm}\left\{\left(\mathbb{Z}/3\mathbb{Z}\right)^{n}\right\}$. Indeed, the update can be defined as:
\[
\begin{array}{l}
p(\mathbf{M}_{i:}\vert \hat{D})\propto p(\hat{D} \vert \mathbf{M}_{i:})\pi(\mathbf{M}_{i:}) = \frac{1}{\Gamma(\alpha)^{m} }\prod_{j=1}^{\bar{n}} \mathbf{M}_{i,j}^{c_{i,j}(\hat{D})}\left(\sum_q\delta_{\mathbf{M}_{i,j}^{(q)}}(\alpha_q-1)\right),\,\text{where,}\\
c_{i,j}(\hat{D}) = \frac{\sum_{l}^{N}\sum_{i,j=1}^{\bar{n},\bar{n}} \mathbf{1}\left((z_{l,t},z_{l,t+1})=(\phi(i),\phi(j))\right)}{\sum_{l}^{N}\sum_{i,j=1}^{\bar{n},\bar{n}} \mathbf{1}\left(z_{l,t}=\phi(i)\right)}
\end{array}
\]
To compute the posterior mixture weights we have,
\[
\begin{array}{l}
p(\mathbf{m}\vert \hat{D})= \int p(\mathbf{m}\vert \mathbf{M})p(\mathbf{M}\vert \hat{D}) d\mathbf{M}=\int p(\mathbf{M}\vert \mathbf{m})\pi(\mathbf{m})p(\mathbf{M}\vert \hat{D}) d\mathbf{M}\\ \qquad \propto \frac{1}{\Gamma(\alpha)^{m}} \prod_{q=1}^{m} [\mathbf{m}]_q^{\alpha_{q}-(\sum_{ij} \mathbb{M}^{(q)}_{ij}-c_{ij})/\bar{n}^2-1}
\end{array}
\]

Now we take the model and obtain the forecast on the training data $\hat{I}=\sum\limits_{i}^I m_i(\hat{D}) \Theta_i^T X_{\mathcal{T}_t}$ and compute the reconstruction loss to the real future in the training $X_{\mathcal{T}_{t+1}}$. It can be see that the closer the empirical frequencies of the training and distilled data are, the smaller the loss. 

Considering a more robust criteria, we can consider formulating an objective that recovers a particular marginal inquiry regarding $\mathbf{M}_{ij}$ accuracy robust with error in the training data. This creates a more complex, bilevel, integer optimization problem.

\section{Case Study Description: DD to construct PGMs for Medical Models}\label{s:dbnmed}
Medical data, broadly speaking, has continued to be analyzed primarily by methods of classic statistics due to the fact that the structure of the data is usually the opposite of the setting amenable to deep learning approaches. Specifically, the number of training samples is generally rather small while the information content of any one sample can be substantial, introducing significant underdeterminacy as far as any attempt to develop an accurate model. It is especially difficult and expensive to establish an experimental and control cohort for clinical trials. Observational studies are more feasible with greater sample size, but often still require inconvenient or even invasive measurement, making recruitment difficult and data collection expensive. By contrast, the rapid growth in the precision of NMR and other technology together with growing biochemical understanding of the human body has created the potential of surveying very detailed \emph{omics} profiles for individuals. In this section, we consider a procedure whereby one can augment a dataset one is working with by systematically constructing additional data from existing data banks and previous research. The problem with simply combining external data directly is often the heterogeneous nature of the choice of features chosen. That is, one dataset can have X and Y, another Y and Z, and yet another Z and X. Imputing missing values can result in significant imprecision and error. However, with DD we can construct a complete dataset that has the most relevant feature information accross disparate datasets. 

Consider that to begin with there are a number of training datasets available $\{\mathcal{T}_n\}$ where each $\mathcal{T}_n$ may have missing data for one or more components. Using this entire collection, the goal is to construct a new dataset $\hat{D}$, for which every sample is complete in information. There are a set $I$ of inference tasks and for each $i$, the map $\iota(i)$ defines the marginal distribution $D(\iota(i))$ associated with that task. For instance, given a person has $a$ background characteristic, and the $b$ covariate is at a value of $4.3$, what is the probability of a particular biomarker for a disease of interest being present. 

We present $O_{PGM(i)}(\cdot)$ as the operator to take a dataset and make a probabilistic graphical model for which the task $i$ is tractable and interpretable. If we collapse the union to just $i$, one evaluation of~\eqref{eq:ddoptex4} would correspond to the PGM built to form a (possibly Bayesian) model $M_{\theta(\omega)}$. This model would perform the $i$-associated inference task on some sample from the distribution and compare it to the ground truth.

We describe our procedure specialized for this problem. At each iteration $k$ we start with the current form of the synthetic dataset $\hat{D}$ and sample uniformly at random some inference task $i$. Each inference task will involve conditioning on some variables, and query on another variable, with the rest as nuisance quantities. A training set will be subsampled from $\{D_n\}$, with a weighting of one for all $D_n$ such that all of the conditioned and query variables are present in $D_n$, and then a fraction weight corresponding to the proportion available for the others. Denote the subsampled training set at $\bar{\mathcal{T}}_k$. A (noisy) evaluation can be performed by learning the model $PGM(i)$ and evaluating the inference query on the model to the query on the subsample. 

Note that in this case there are a number of methods to train a PGM, many of which without gradient trajectories, and also that without a neural network other approaches, such as based on NTK, are not appropriate. At the same time, a gradient of the original problem is not analytically available. We propose a zero order approximation to the gradient, using a classic two point function approximation rule. 

\paragraph{Zero Order Optimization}

We formally detail the first procedure. Algorithm~\ref{alg:eval} describes an evaluation of the objective of interest with respect to the input decision variable, the synthetic dataset $\hat{D}$. 

In this algorithm, which requires an inference task $i$ and subsample $\bar{\mathcal{T}}$, the first step is to train a Probabilistic Graphical Model on the synthetic dataset. This returns a model defined as $M_{\theta(\omega)}$. In the numerical experiments, we use Dynamic Bayesian Networks, DBNs. 

Subsequently, we evaluate the model $M_{\theta(\omega)}$ on the inference task as performed on the training subsample $\bar{\mathcal{T}}$. Finally, a probability distance is computed between the computed inference task $\hat{I}$ and the inference task ground truth from some validation set $I(D_{(i)})$. Some statistical operation across the training subsample is performed, whether an expectation, or a risk measure like Conditional Value at Risk (CVaR). 

\begin{algorithm}[H]
\caption{DD Loss Evaluation}\label{alg:eval}
\textbf{Input:} Current synthetic dataset $\hat{D}$. Inference task $i\in I$. Subsample $\bar{\mathcal{T}}$, hold out validation set $\mathcal{V}$
\begin{algorithmic}[1]
\State Compute $M_{\theta(\omega)}=O_{PGM(i)}(\hat{D})$
\State Compute $\hat{I}=\mathcal{I}(M_{\theta(\omega)},\bar{\mathcal{T}}_{\iota_i})$ using EM to impute missing data, where $\bar{\mathcal{T}}_{\iota_i}$ refers to the training data relevant to the calculation of the inference task. 
\State Return the value of,
\[
\mathcal{R}_{\mathcal{T}_{\iota(i)}} \left[\mathcal{D}_{\theta(\omega)}\left(\hat{I} \vert \vert I(\mathcal{V}_{(i)}) \right)\right]
\]
where $I(\mathcal{V}_{(i)})$ defines the inference procedure on the validation set. 
\end{algorithmic}
\end{algorithm}

In order to compute~\eqref{eq:gest} we can consider two classes procedures at either end of the variance accuracy tradeoff spectrum in zero order or derivative free optimization (see, \textit{e.g.}~\cite{berahas2022theoretical}).  

The first, based on sparse noisy function evaluations, is common in machine learning. The \emph{Two Point Function Approximation} Estimate is given by:
\begin{equation}\label{eq:2pfa}
 g_k= \frac{1}{M}\sum\limits_{l=1}^M\frac{L(\hat{D}+\sigma v_l,i)-L(\hat{D}_k,i)}{\sigma}v_l
\end{equation}
with $v_l\sim \mathcal{N}(0,I)$, where $L(\hat{D})$ is the evaluation given by Algorithm~\ref{alg:eval} with inputs $\hat{D}$ and $i$.

\paragraph{First Order Algorithm}

The actual optimization procedure is defined in Algorithm~\ref{alg:opt}
\begin{algorithm}[H]
\caption{Optimization Loop}\label{alg:opt}
\textbf{Input:} Training set $\mathcal{T}$ and set of inference tasks $I$
\begin{algorithmic}[1]
\For{$k=0,1,2,...$}
\State Sample $i\in I$
\State Subsample, with importance weighing on datasets without missing features,  $\bar{\mathcal{T}}_k\subset \mathcal{T}$. 
\State Compute an estimate of the gradient of $L$ with respect to the synthetic dataset:
    \begin{equation}\label{eq:gest}
    g_k\approx \nabla L(\hat{D}_k) 
    \end{equation}
    
    \State Set stepsize $s$ (diminishing stepsize rule, or stepwise decay). For instance:
    \[
    s=0.1/\sqrt{1+k}
    \]
    or a stepwise decay
    \State  Set $\hat{D}_k-s g_k\to \hat{D}_{k+1}$
\EndFor
\end{algorithmic}
\end{algorithm}

Note that alternatively, when the dimensionality isn't prohibitive, a more accurate and less noisy estimate can be performed, such as the one in commonly used in Derivative Free Optimization (DFO) with schemes to obtain accuracy with high probability~\cite{larson2016stochastic}. We found the performance of the zero order approximator was superior in this case, however. 

To illustrate the proof of concept we will perform the evaluation of the inference task of a test likelihood. That is $\mathcal{I}$ is the total likelihood and $\mathcal{R}$ is taken to be 

\subsection{Experimental Setup}

\begin{figure}[t]
    \centering
    \vspace{-8pt}
    \includegraphics[scale=0.3]{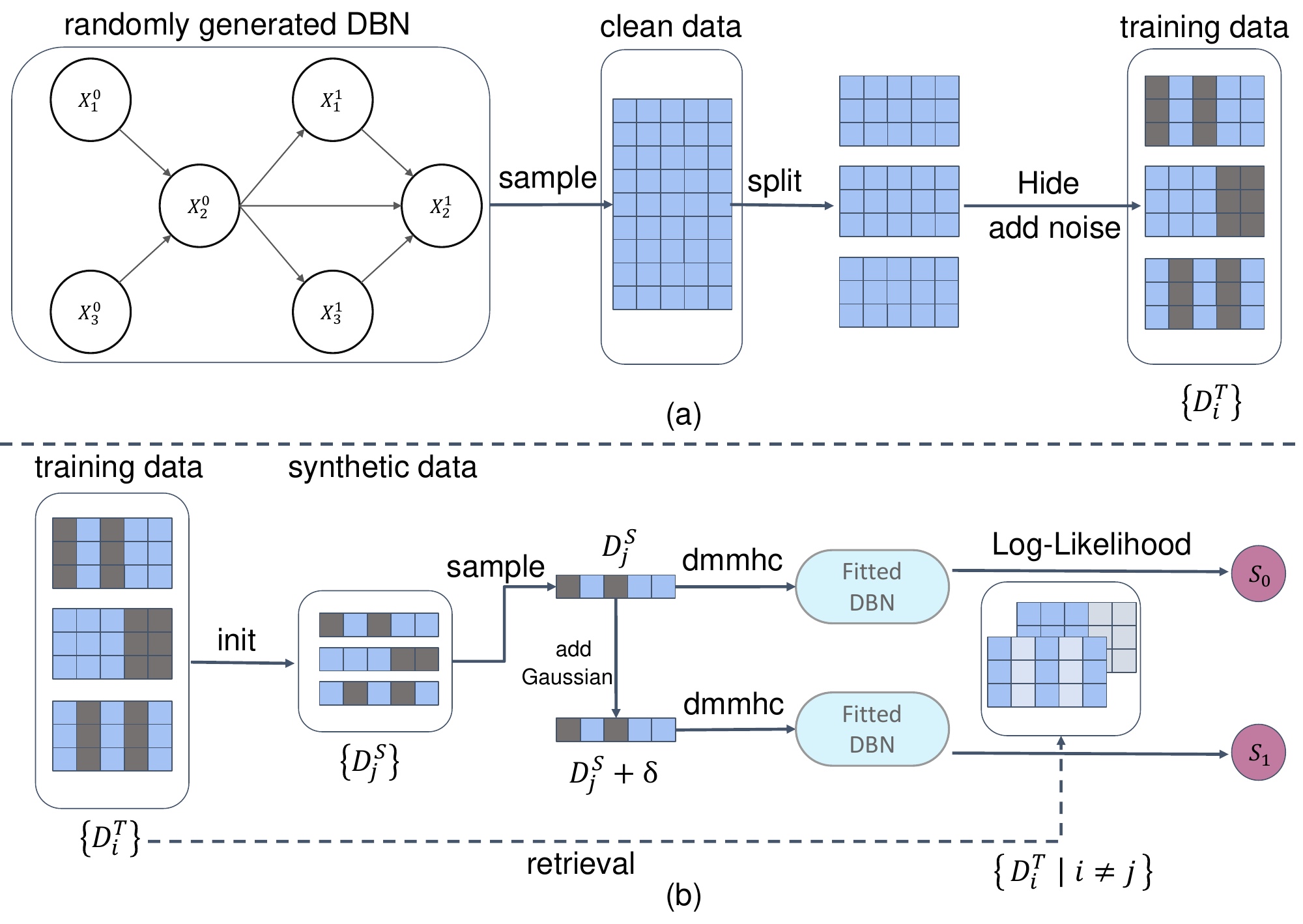}
    \caption{(a) We generate datasets by sampling from a randomly initailized DBN. To mimic heterogeneity of medical data, we split the sampled clean data to $K$ partitions. For each partition, we hide the values of different variables and add Gaussian noise to each entry. (b) To distill the synthetic data more effectively, we initialize $D^{S}$ from each source of training data. During each epoch, we sample a part $D_{j}^{S}$ from $D^{S}$, add Gaussian noise $\delta$ to it, learn two different DBNs from them respectively. Then retrieve training data $\left\{ D_i^T \mid i \ne j \right\}$. Finally we compute the log-likelihood of $D_{j}^{S}$ in two different DBNs as the evaluation score, to calculate the gradient with respect to  $D_{j}^{S}$. }
    \lblfig{pipeline}
         \vspace{-8pt}
\end{figure}

\paragraph{Dataset:}
As illustrated in \reffig{pipeline}, we use a randomly generated Gaussian Dynamic Bayesian Network (GDBN) to generate training and testing datasets. Source nodes (indegree equal to 0) are Gaussian nodes with a mean value of 0 and a standard deviation of 0.5 . The mean values of intermediate nodes and target nodes are linear combinations of parent nodes, while the standard deviation is also set to 0.5.  For computational efficiency, the GDBN consists of 20 variables and 2 time-slices. We sample 1000 observations from the GDBN as the testing dataset $D^{Test}$ . For training dataset, we firstly sample 1000 observations, then split them to 5 partitions. For each partition, 10\% of the variables are randomly hidden, and Gaussian noise is added independently, to mimic the heterogeneity of medical datasets from different data sources. We also investigate the performance of our algorithm when scaling up time slices. Specifically, we test our algorithm in 20 variables and 10 time-slices setting\protect\footnotemark, keeping other experimental settings the same.

\footnotetext{The experiment of scaling to more variables and time slices is prohibitively time-consuming because of bi-level optimization nature of DD and structure learning algorithm $O_{PGM}(\hat{D})$, then we only scale time-slices from 2 to 10 in this paper.}

\begin{figure}[ht]
    \centering
    \vspace{-10pt}
    \includegraphics[scale=0.4]{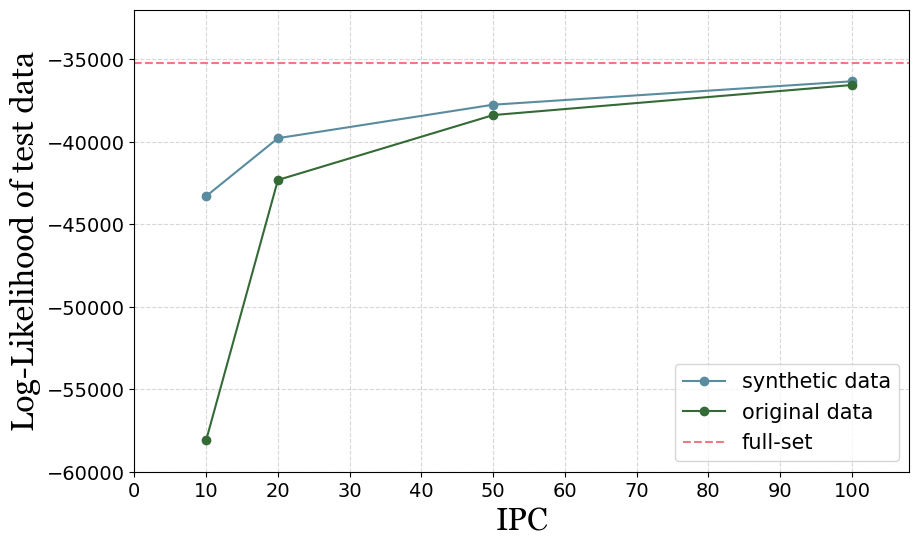}
    \caption{Performance of our algorithm in 4 different IPC settings. Across 4 settings, our algorithm is much better than fully-observed clean data $D^{sub}$, especially in low-data regime. When IPC=100, our algorithm is almost as effective as the complete and clean training dataset.}
    \lblfig{main_result}
         \vspace{-10pt}
\end{figure}

\begin{figure}[h]
    \centering
    \begin{tabular}{cc}
        \begin{subfigure}[b]{0.45\linewidth}
            \centering
            \includegraphics[width=\linewidth]{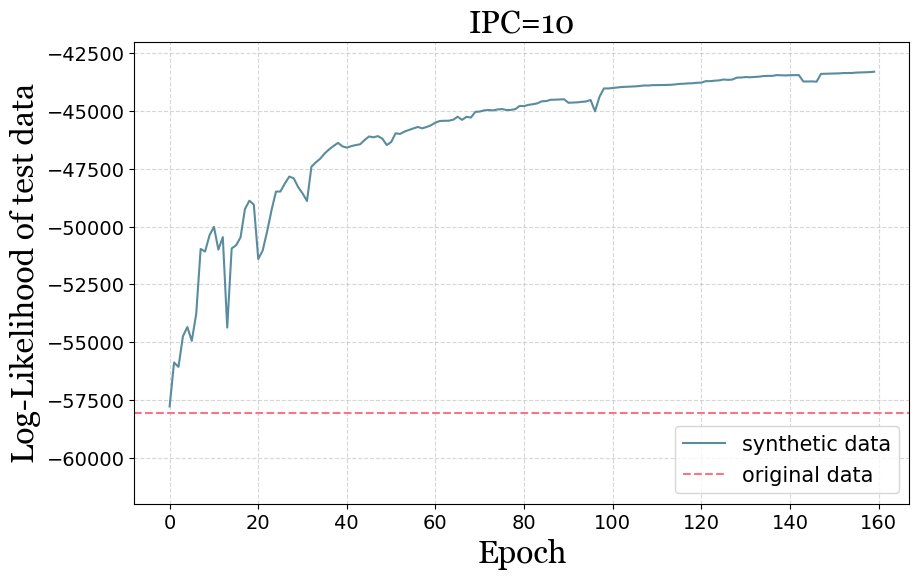}
            \caption{}
            \label{curve1}
        \end{subfigure} &
        \begin{subfigure}[b]{0.45\linewidth}
            \centering
            \includegraphics[width=\linewidth]{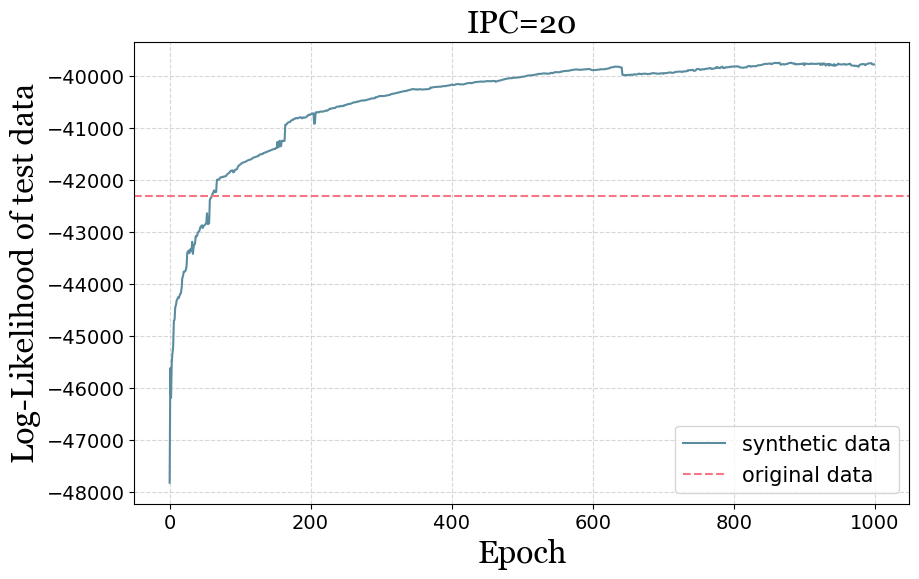}
            \caption{}
            \label{curve2}
        \end{subfigure} \\
        \begin{subfigure}[b]{0.45\linewidth}
            \centering
            \includegraphics[width=\linewidth]{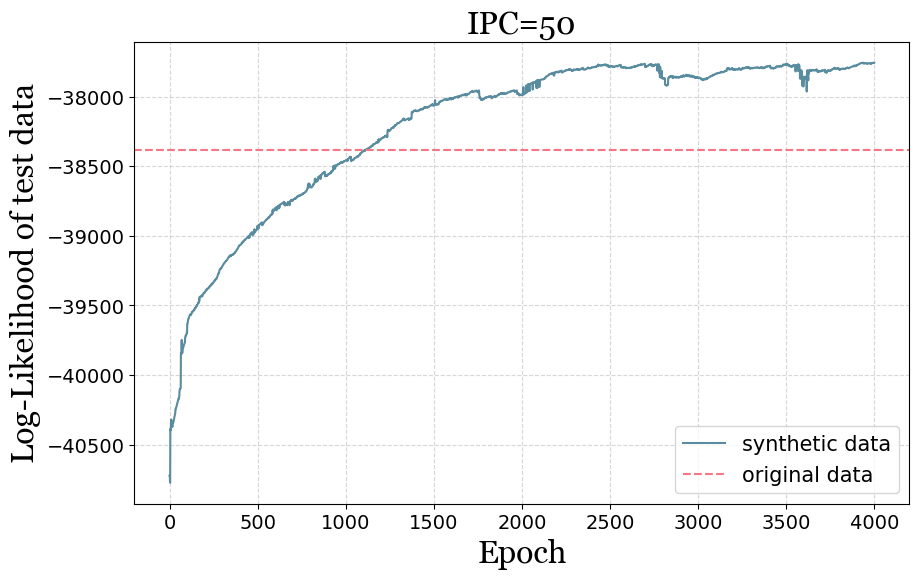}
            \caption{}
            \label{curve3}
        \end{subfigure} &
        \begin{subfigure}[b]{0.45\linewidth}
            \centering
            \includegraphics[width=\linewidth]{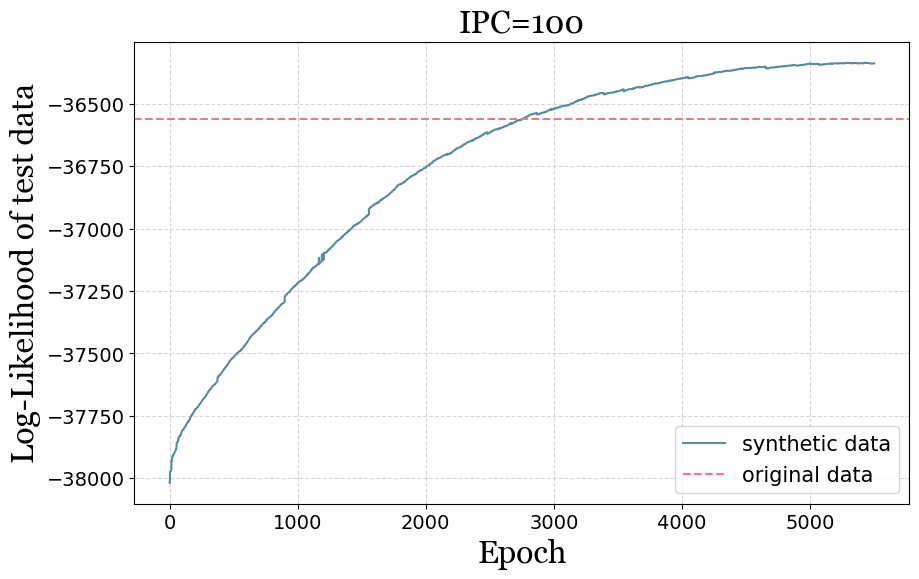}
            \caption{}
            \label{curve4}
        \end{subfigure}
    \end{tabular}
    \caption{The Log-Likelihood (LL) curves in testing set $D^{Test}$ of 4 different IPC settings. We evaluate the synthetic dataset $\hat{D}$ after each distillation process and report LL, with comparison to $D^{sub}$}
    \lblfig{LL_curve}
\end{figure}

\paragraph{Evaluation and Baseline:}
After distilling the synthetic dataset $\hat{D}$, we use $O_{PGM}(\hat{D})$ to fit a GDBN $M_{\theta(\omega)}$. The Log-Likelihood of  $D^{Test}$ on $M_{\theta(\omega)}$ is computed as the evaluation metric. For comparison, we randomly sample the same quantity of fully-observed clean observations $D^{sub}$ from training dataset before hiding variables and add noise to it. It is a challenging baseline because $D^{sub}$ is fully-observed with no noise, while our algorithm starts with partially observed dataset with Gaussian noise. We use IPC to denote quantity of observations in synthetic dataset and subset of training dataset in baseline. For both our algorithm and baseline, we investigate 4 
 IPC settings: 10, 20, 50, 100 . We also test the performance of the complete fully-observaed clean training set as the upper bound.

\paragraph{Implementation Details}
For approximating the gradient of objective with respect to $\hat{D}$ with lower variance, we set $M$ as 10 in~\eqref{eq:2pfa}. Optimization iterations is set to 160, 1000, 4000, 5500 for IPC 10, 20, 50, 100 respectively. The learning rate is kept as 4e-5 along the distillation process without scheduling and special optimizer. We initialize $\hat{D}$ with 5 sources of training dataset, with missing values sampled from Gaussian, for we observed that it converge faster and performs better than purely initializing $\hat{D}$ from Gaussian. The $O_{PGM}(\hat{D})$ operator and LL computation are adopted from dbnR in~\cite{Quesada2024dbnR}.

\subsection{Numerical Results }

\paragraph{Main Results:}
We compare our algorithm with baseline in 4 different IPC settings. As illustrated in \reffig{main_result}, across 4 settings, our algorithm illustrates significant superiority over even the fully-observed clean data $D^{sub}$, especially in low-data regime. When IPC=100, our algorithm is almost as effective as the complete and clean training dataset. However, when IPC gets larger, the marginal rate of performance increase of both our algorithm and $D^{sub}$ declines. This is like to be because of the growing redundancy in $\hat{D}$ and $D^{sub}$.
The training curves of different IPC settings during the distillation process are shown in \reffig{LL_curve}. Although faced with Gaussian noise, missing values in training data and the black-box optimization nature of the algorithm, the training curves show steady gradient and performance growth and outperform the strong baseline across 4 different settings.

\paragraph{Scaling Up Time-slices}
Medical data typically has many time-slices. To test the scalability of our algorithm, we scale time-slices from 2 to 10, keeping other settings the same. As shown in \reffig{scaling_result}, the synthetic data is again even better than the fully-observed clean data across all settings. From IPC=20 to larger quantities of synthetic data, its performance is very close to original full training dataset, indicating the consistent scalability of our algorithm. However, we observed that time consumption of DD grows linear with time-slices, which is intuitive because the outer loop evaluation has to compute values of missing variables and calculate LL slice by slice. How to scale to even larger problem dimension, sample size, and time-slices is beyond the scope of this work and can be tackled in the future.

\begin{figure}[t]
    \centering
    \vspace{-8pt}
    \includegraphics[scale=0.4]{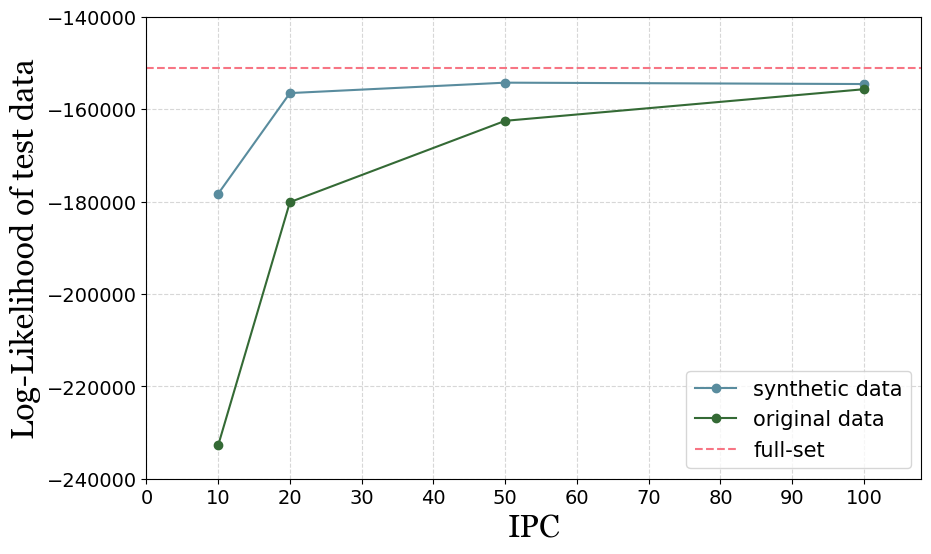}
    \caption{Performance of our algorithm when scaling up time-slices from 2 to 10. Across 4 IPC settings, our algorithm is much better than baseline.}
    \lblfig{scaling_result}
         \vspace{-8pt}
\end{figure}

\section{Case Study Description: DD to Construct Boundary-Generalizable PINNs}\label{s:phys}

\begin{figure}[t]
    \centering
    \vspace{-16pt}
    \includegraphics[scale=0.4]{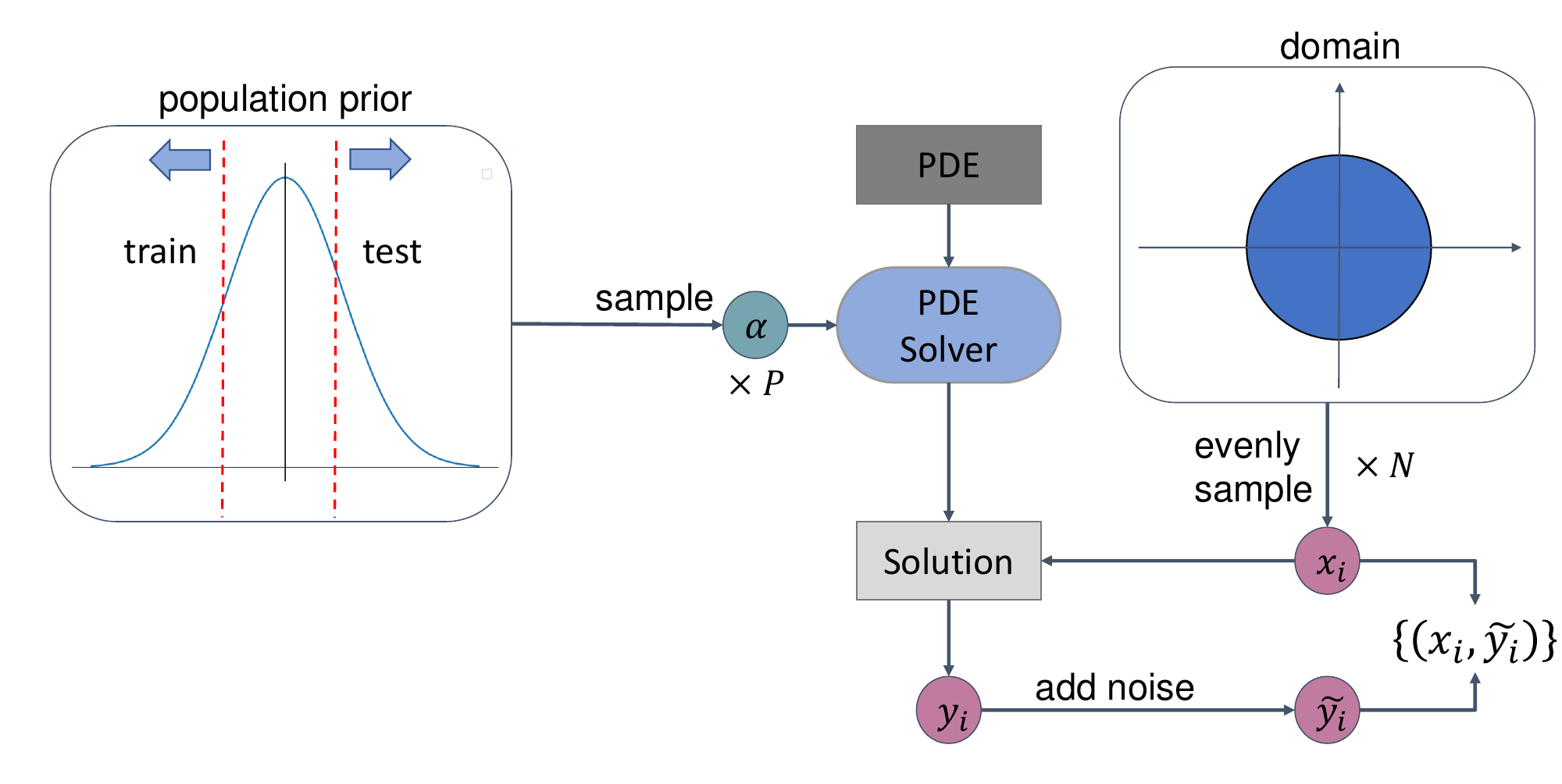}
    \caption{Training and test dataset generation of PINN example. To demonstrate the out-of-distribution generalization ability of our algorithm, boundary conditions ($\alpha$) of training and test dataset are sampled from two opposite tails of normal distribution. For training dataset, noise is added to ground truth $y_i$ to mimic measurement error, while for test dataset, none noise is added to $y_i$}
    \lblfig{data_pinn}
         \vspace{-8pt}
\end{figure}

Recall the PDE loss function given in Section~\ref{s:ex}
\[
\mathcal{L}_F(\theta_u,\theta_{\gamma};\{x\},\hat{f})=\frac{1}{MN}
\sum\limits_{m=1}^M\sum\limits_{l=1}^N (\mathcal{F}(\bar{u}_l(\theta,x_m),\bar{\gamma}_l(\theta,x_m))-\hat{f}_{ml})^2
\]
and the bilevel optimization problem defining the DD procedure:
\[
    \begin{array}{rl}
    \min\limits_{\hat{D},\theta} & \mathbf{CVaR}_{\pi\left(u_{\partial\Omega}\right)} \left[\mathcal{L}_F(\theta_u,\theta_{\gamma};D)\right]\\ 
    \text{subject to } & \theta = \arg\min \mathcal{L}_F(\theta_u,\theta_{\gamma};\hat{D}) \\ 
    & \qquad \text{s.t. } u(x_{\partial\Omega},\theta,g) = g(x_{\partial\Omega}) \sim \hat{D}_g 
    \end{array}.
\]
We further design corresponding experiments on PINN to demonstrate the effectiveness of our proposed dataset distillation scheme. 

\subsection{Experimental Setup}
\paragraph{Dataset}
We deal with Laplace equation on a disk with Dirichlet boundary conditions:
\[
r \frac{dy}{dr} + r^2 \frac{d^2 y}{dr^2} + \frac{d^2 y}{d\theta^2} = 0, \quad r \in [0, 1], \quad \theta \in [0, 2\pi]
\]
\[
y(1, \theta) = \cos(\theta)+\alpha
\]
\[
y(r,\theta+2\pi)=y(r,\theta)
\]
As demonstrated in \reffig{data_pinn}, we sample boundary conditions $\alpha$ from two opposite tails of population prior $\mathcal{N}(0,1)$, to verify the out of distribution (OOD) generalization ability of our algorithm. Specifically, $\alpha_{train}$ is restricted to be less than $Q(a)$ and $\alpha_{test}$ is restricted to be greater than $Q(1-a)$, where $Q(a)$ is quintile function of $\mathcal{N}(0,1)$. Then a numerical PDE solver is used to solve the PDE equation with $\alpha$ and return a solution\protect\footnotemark $y=f(x)$. Finally we evenly sample from a lattice of points ${x_i}$ in the domain and obtain the ground truth ${y_i}$ by $y=f(x)$. For the training dataset, Gaussian noise is added to ${y_i}$ to mimic the measurement error, while for the test dataset, no noise is added.  $P$ in \reffig{data_pinn} is set to 100. $N$ is set to 2620, where 2540 points inside the domain and 80 points on the boundary. Noise is sampled from $N(0,0.05)$. $a$ is set to $40\%$.

\footnotetext{For the Laplace equation on a disk with Dirichlet boundary conditions, the analytic solution is known: $y=r\cos{\theta}+\alpha$.}

\paragraph{Implementation Details}
The implementation is based on DeepXDE in~\cite{lu2021deepxde}. When learning a PINN from a synthetic dataset, we use a 4-layer FCN as the network with $tanh$ activations. The loss function consists of prediction loss $l(y_i,\hat{y_i})$ and the PDE-constrain loss $\vert \Delta y_i + \pi^2 \sin(\pi x_i) \vert$. The learning rate and distilling epoch number are set to 0.001 and 1000, respectively, with an Adam optimizer adopted. When optimizing  the synthetic dataset, we  adopt DFO algorithm NgIohTuned in Nevergrad ~\cite{nevergrad} to search for the best values of synthetic dataset. For the only significant parameter $budget$ in NgIohTuned, we experiment with 5, 10, 20, 40,100 and 200 to find the best value. For evaluation, we learn a PINN from the synthetic dataset as described earlier and report the L2 error on test dataset. We compare our synthetic dataset with random sampled subsample from the training dataset without noise of the same quantity. We investigate 4 IPC settings: 10, 20, 40 and 80. We initialize the synthetic dataset from Gaussian noise when IPC=10 and from randomly sampled data from training dataset when IPC=20,40,80 for efficacy-efficiency trade-off.

\subsection{Numerical Results}

\begin{table}[t]
    \centering
    \scriptsize
    \resizebox{0.6\linewidth}{!}{
    \begin{tabular}{cc|cccc}
        \toprule
        & \multirow{2}{*}{Budget} &\multicolumn{4}{c}{IPC}\\
        & & 10 & 20 & 40 & 80  \\
        \midrule \multirow{4}{*}{\rotatebox[origin=c]{90}{Ours}} &5 & 0.1310 & 0.1056 & 0.1288 &0.1419\\
        &10 & 0.1752 & 0.0995 & 0.0595 &  0.0726\\
        &20 & \textbf{0.1258} & 0.1248&  \textbf{0.0346}& \textbf{0.0186}\\
        &40 &0.8427 &0.0973 &0.1794 &0.0320\\
        &100 &2.1470 &\textbf{0.0602} &0.0417 &0.0307 \\
        &200 &0.8130 &0.1609 &0.0801 &0.0506 \\
        \midrule
        \multicolumn{2}{c|}{Subset}&0.2003 &0.1167 &0.1008 &0.0552\\
        \bottomrule
    \end{tabular}
    }
    \caption{Comparison of our algorithm with baseline in PINN example. Across 4 IPC settings, ours has smaller prediction error than baseline.}
    \lbltbl{pinn_result}
\end{table}

We investigate our algorithm and compare with baseline in 4 IPC settings. For each IPC setting, we try 6 budgets to find the best budget of the DFO algorithm. As shown in \reftbl{pinn_result}, our algorithm outperforms baseline in all the 4 different IPC settings. Although faced with noise, our algorithm shows smaller prediction error and stronger OOD generalization ability. When IPC is equal to 80, the generalization error is only one quarter of the strong baseline.

\paragraph{PDE Reconstruction Visualization}
To visualize PDE reconstructed solution of our method and baseline, we sample a boundary condition $\alpha$ from test-side tail in \reffig{data_pinn} and reconstruct the PDE with $\alpha$ boundary condition by training a PINN on our synthetic dataset and baseline dataset respectively. As shown in \reffig{pinn_vis}, both our method and baseline can roughly reconstruct the outline of ground truth solution, while our method shows much smaller reconstruction error compared with baseline, which verifies the effectiveness of our method.

\begin{figure}[t]
    \centering
    \begin{tabular}{ccc}
        \begin{subfigure}[b]{0.35\linewidth}
            \centering
            \includegraphics[width=\linewidth]{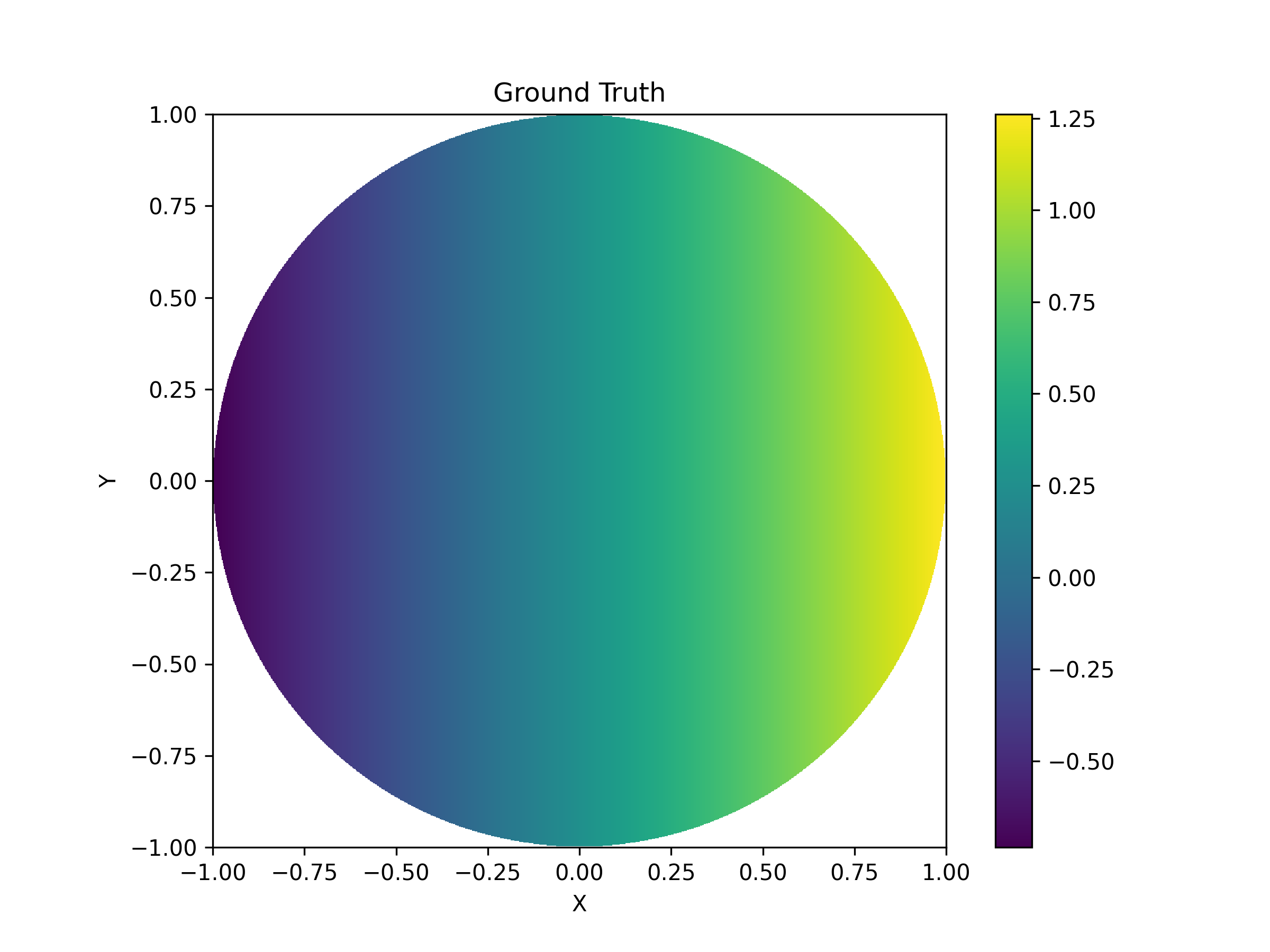}
            \caption{}
            \label{curve1_pinn}
        \end{subfigure} &
        \begin{subfigure}[b]{0.35\linewidth}
            \centering
            \includegraphics[width=\linewidth]{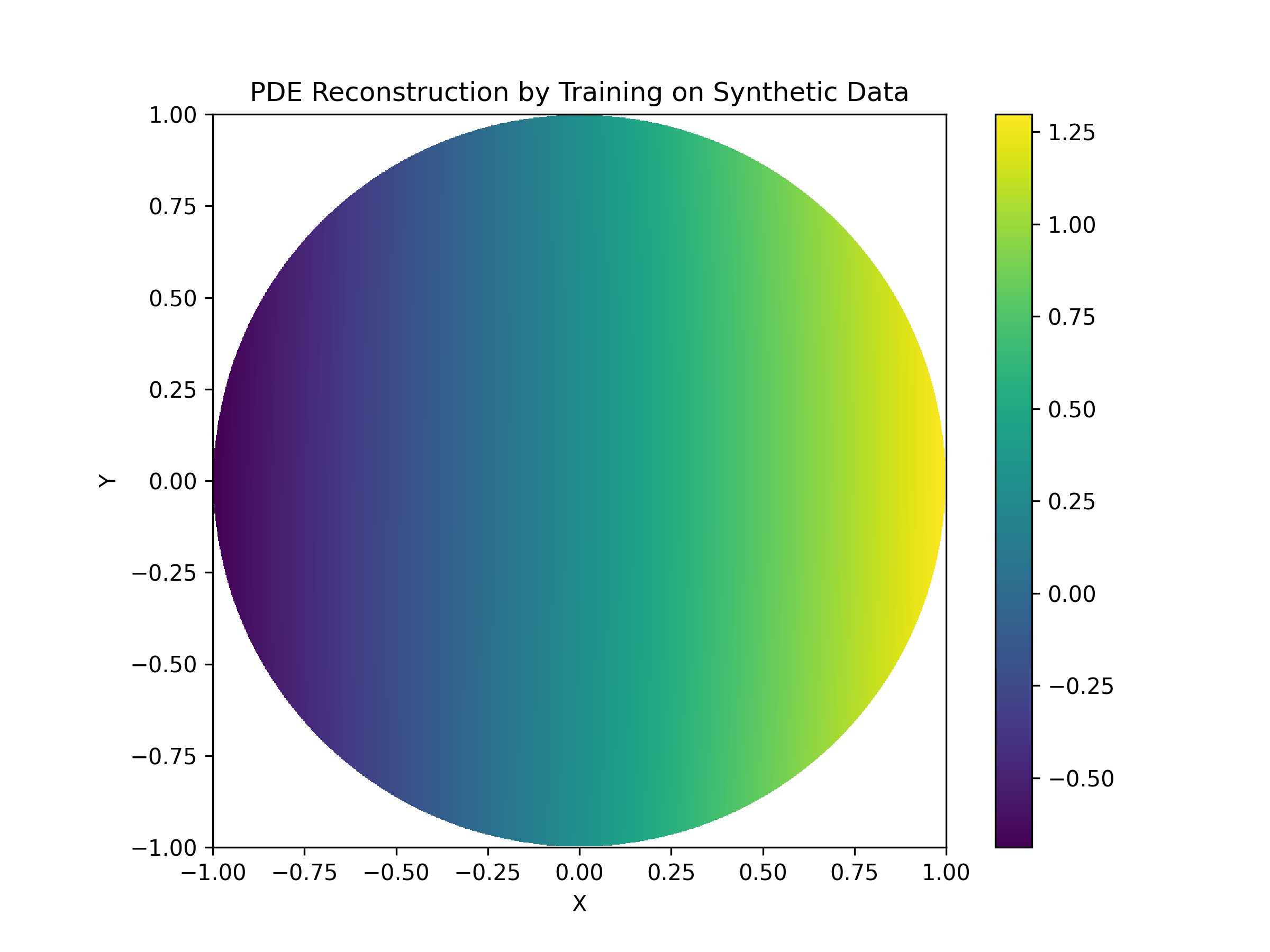}
            \caption{}
            \label{curve2_pinn}
        \end{subfigure} &
        \begin{subfigure}[b]{0.35\linewidth}
            \centering
            \includegraphics[width=\linewidth]{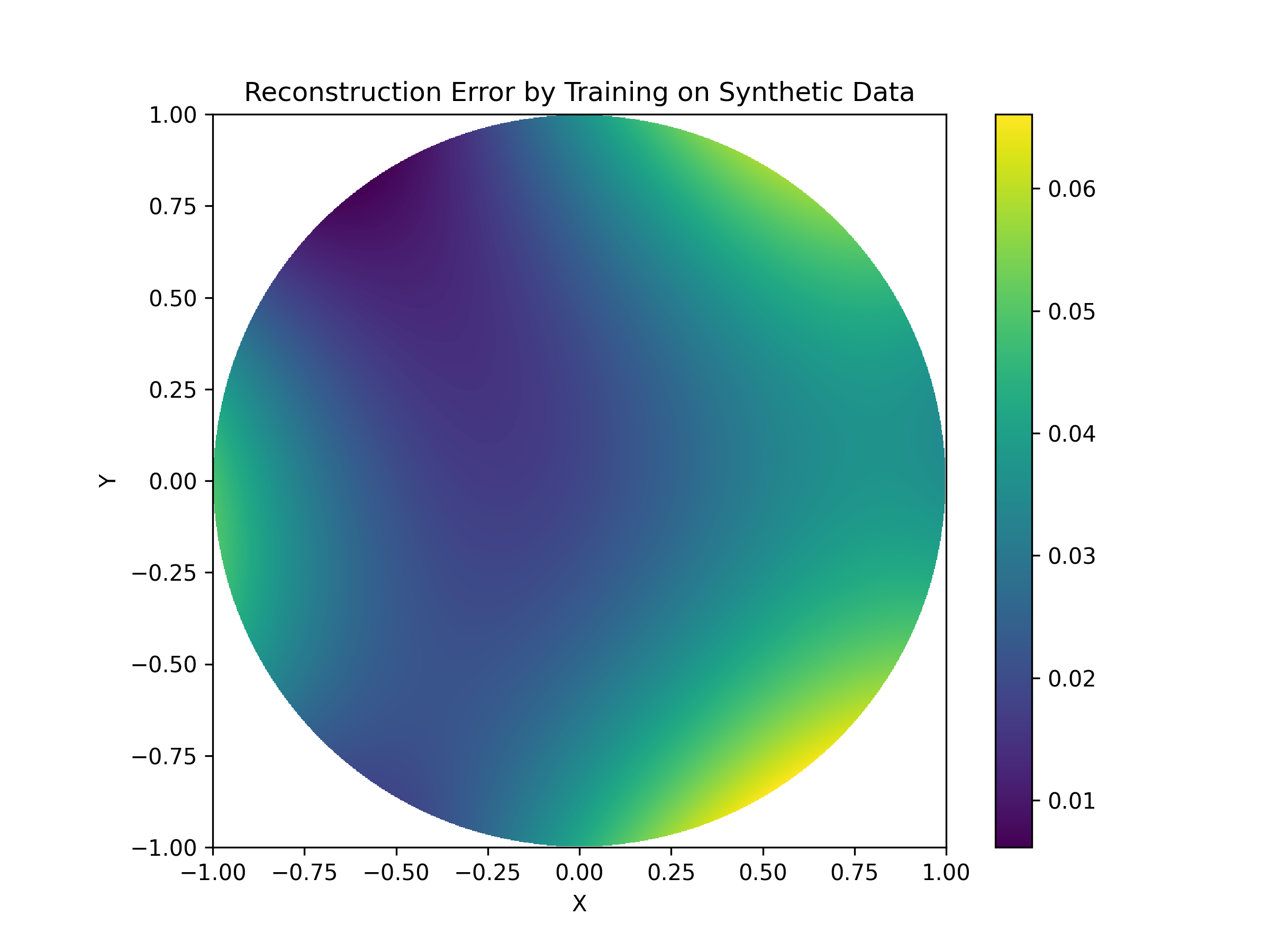}
            \caption{}
            \label{curve3_pinn}
        \end{subfigure} \\
        \begin{subfigure}[b]{0.35\linewidth}
            \centering
            \includegraphics[width=\linewidth]{figures/GT_pinn.png}
            \caption{}
            \label{curve4_pinn}
        \end{subfigure}&
        \begin{subfigure}[b]{0.35\linewidth}
            \centering
            \includegraphics[width=\linewidth]{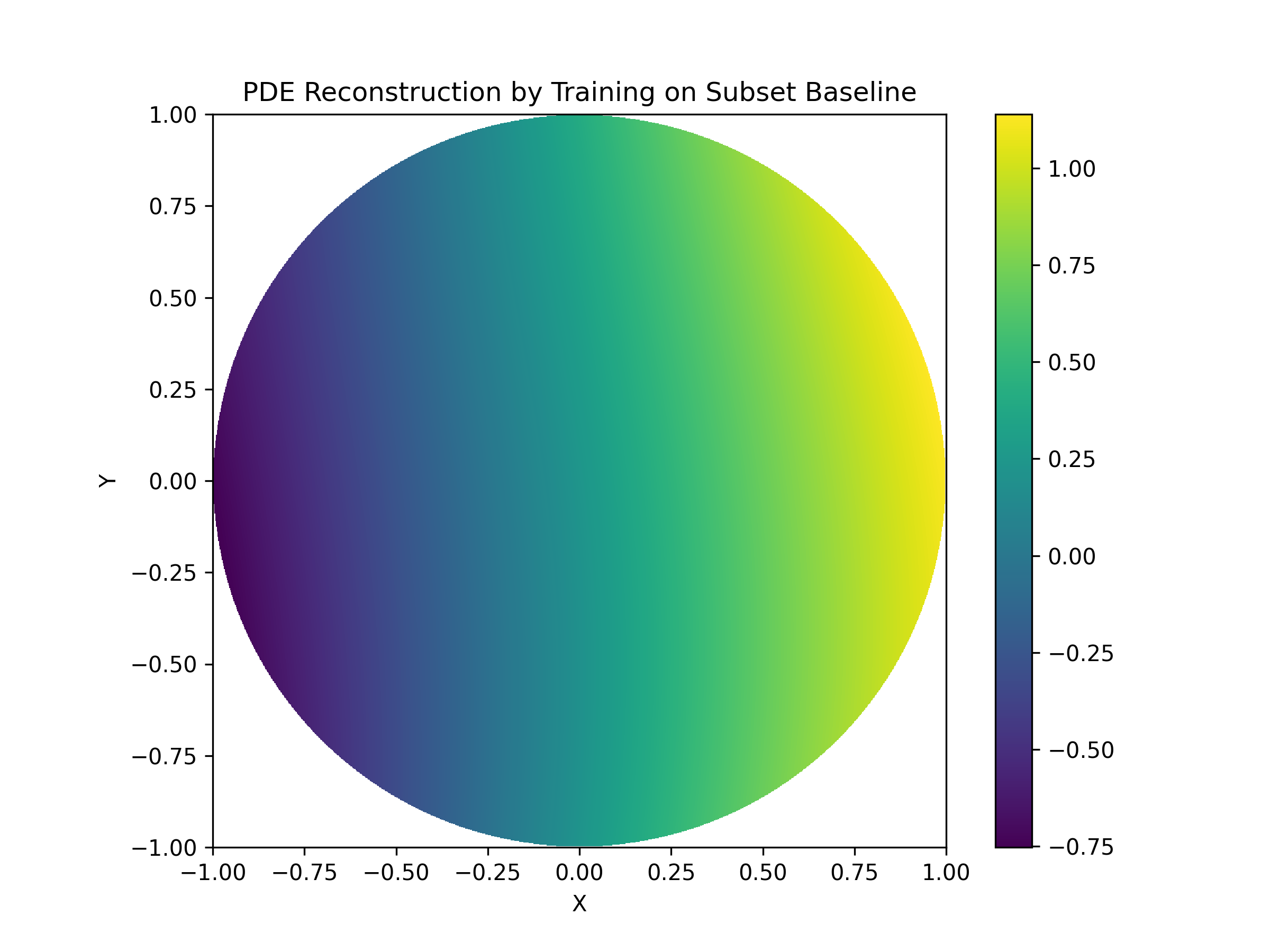}
            \caption{}
            \label{curve5_pinn}
        \end{subfigure}&
        \begin{subfigure}[b]{0.35\linewidth}
            \centering
            \includegraphics[width=\linewidth]{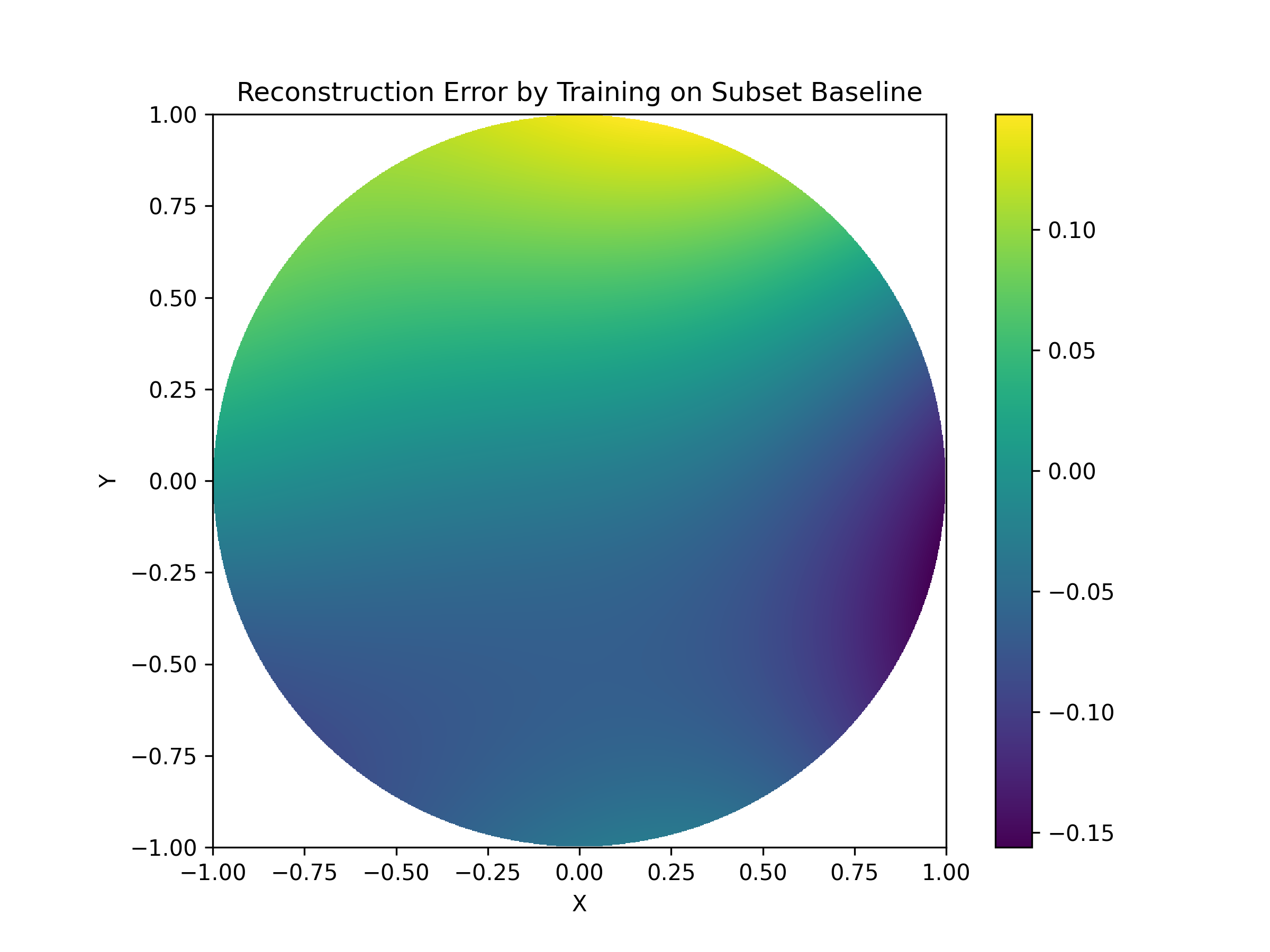}
            \caption{}
            \label{curve6_pinn}
        \end{subfigure}
    \end{tabular}
    \caption{PDE reconstruction visualization. $(x,y)$ is the coordinate in the 2D domain. Value of the function or error is represented by color. (a) and (d) are ground truth of PDE solution. (b) and (e) are PDE reconstructed solution given by PINN trained on our synthetic data and baseline respectively. (c) and (f) show the reconstruction error of our method and baseline}
    \lblfig{pinn_vis}
\end{figure}

\begin{figure}[h]
    \centering
    \includegraphics[scale=0.35]{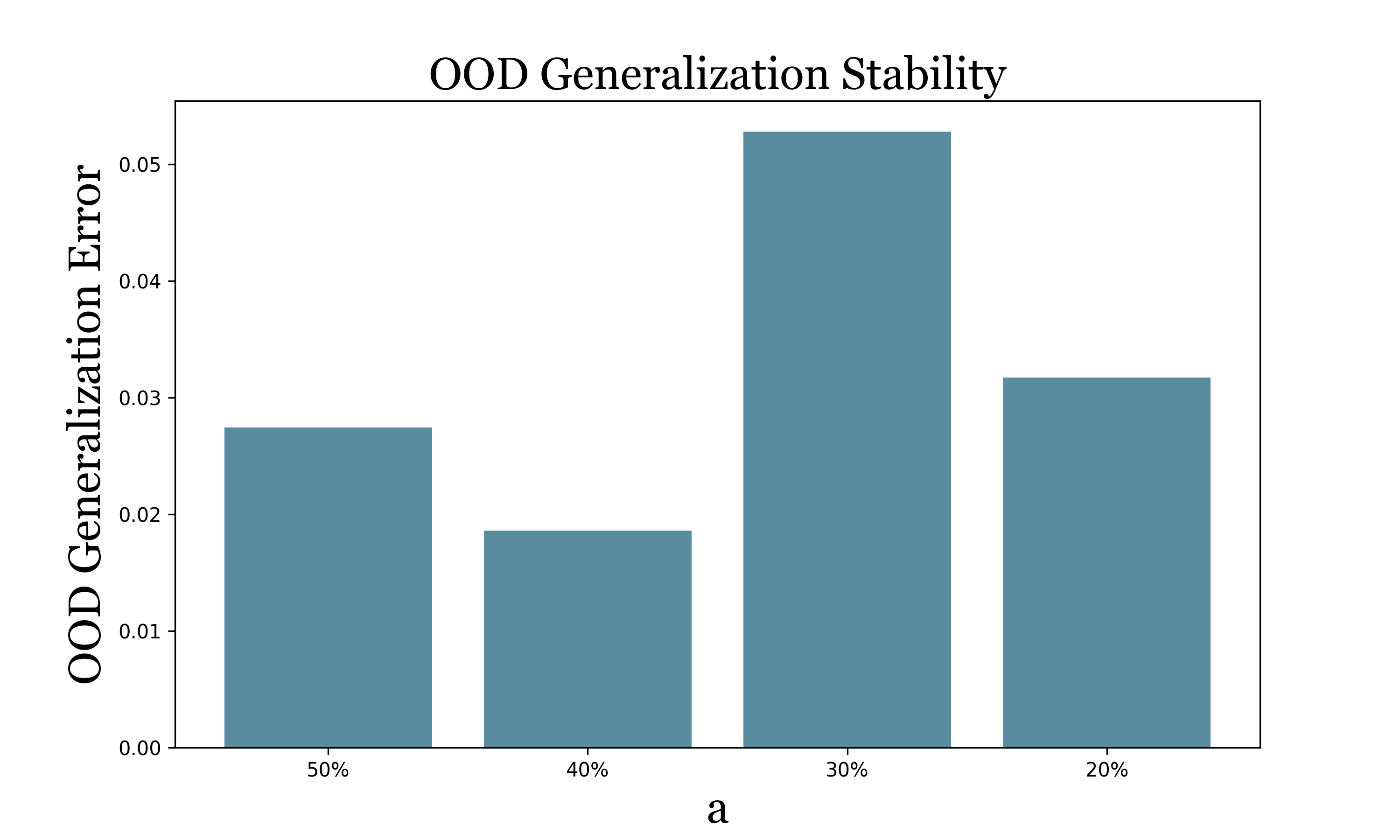}
    \caption{Experimental result of OOD generalization stability. The horizontal axis $a$ is parameter of quintile function $Q(a)$, which is varied from $50\%$ to $20\%$ to control the distance between training population prior and testing population prior.}
    \lblfig{ood_general}
\end{figure}

\paragraph{OOD Generalization Stability}
To test the OOD generalization stability, the distance of two tails as shown in \reffig{data_pinn} is varied. In other words, the parameter $a$ of quintile function $Q(a)$ is set to $50\%,40\%,30\%,20\%$ and PDE reconstruction error of test dataset is computed as OOD generalization error under these 4 settings. As seen in \reffig{ood_general}, our algorithm shows strong OOD generalization stability when $a$ varies from $50\%$ to $20\%$. The OOD generalization error could be restricted to smaller than or around 0.05.

\section{Conclusion}\label{s:conc}
In this paper we provided a general formalization of DD, and presented a broad theoretical understanding of the optimization problem so defined. With this formalization we can define a number of interesting existing and new applications of DD, as well as provide an interpretive framework by which to design algorithms. We analyzed existing procedures for DD and provided a set of illustrative examples of novel applications. We hope that this work serves to provide useful scaffolding to assist future research activity in insight into DD and new DD methods.

\bibliography{refs.bib}
\end{document}